\pdfoutput=1
\documentclass[11pt,a4paper]{article}
\usepackage[hyperref]{emnlp2020}
\usepackage{times}
\usepackage{latexsym}

\usepackage{microtype}

\aclfinalcopy 
\def\aclpaperid{1983} 

\usepackage{booktabs} 
\usepackage{multirow}

\usepackage{graphics}
\usepackage{graphicx}
\usepackage{caption, subcaption}

\usepackage{xcolor} 

\definecolor{bert-orange}{rgb}{1.0, 0.498, 0.054}
\definecolor{roberta-blue}{rgb}{0.121, 0.0.466, 0.705}
\definecolor{albert-green}{rgb}{0.172, 0.627, 0.172}

\usepackage{amsmath}
\usepackage{siunitx}
\title{On the Interplay Between Fine-tuning and Sentence-level Probing for Linguistic Knowledge in Pre-trained  Transformers}

\author{Marius Mosbach \quad Anna Khokhlova \quad Michael A. Hedderich \quad Dietrich Klakow \\
  Spoken Language Systems (LSV) \\
  Department of Language Science and Technology \\
   Saarland Informatics Campus, Saarland University, Germany \\
  \texttt{\{mmosbach,akhokhlova,mhedderich,dklakow\}@lsv.uni-saarland.de}
  }

\date{}

\begin{document}
\maketitle

\begin{abstract}

Fine-tuning pre-trained contextualized embedding models has become an integral part of the NLP pipeline. At the same time, probing has emerged as a way to investigate the linguistic knowledge captured by pre-trained models. Very little is, however, understood about how fine-tuning affects the representations of pre-trained models and thereby the linguistic knowledge they encode. This paper contributes towards closing this gap. We study three different pre-trained models: BERT, RoBERTa, and ALBERT, and investigate through sentence-level probing how fine-tuning affects their representations. We find that for some probing tasks fine-tuning leads to substantial changes in accuracy, possibly suggesting that fine-tuning introduces or even removes linguistic knowledge from a pre-trained model.
These changes, however, vary greatly across different models, fine-tuning and probing tasks. Our analysis reveals that while fine-tuning indeed changes the representations of a pre-trained model and these changes are typically larger for higher layers, only in very few cases, fine-tuning has a positive effect on probing accuracy that is larger than just using the pre-trained model with a strong pooling method. Based on our findings, we argue that both positive and negative effects of fine-tuning on probing require a careful interpretation. 

\end{abstract}

\section{Introduction}

Transformer-based contextual embeddings like BERT \cite{devlin-etal-2019-bert}, RoBERTa \cite{liu2019roberta} and ALBERT \cite{Lan2020ALBERT:} recently became the state-of-the-art on a variety of NLP downstream tasks. These models are pre-trained on large amounts of text and subsequently fine-tuned on task-specific, supervised downstream tasks. Their strong empirical performance triggered questions concerning the linguistic knowledge they encode in their representations and how it is affected by the training objective and model architecture \cite{kim-etal-2019-probing, wang-etal-2019-tell}. One prominent technique to gain insights about the linguistic knowledge encoded in pre-trained models is \textit{probing} \cite{rogers2020primer}. However, works on probing have so far focused mostly on pre-trained models. It is still unclear how the representations of a pre-trained model change when fine-tuning on a downstream task. Further, little is known about whether and to what extent this process adds or removes linguistic knowledge from a pre-trained model. Addressing these issues, we are investigating the following questions: 

\begin{enumerate}
    \item How and where does fine-tuning affect the representations of a pre-trained model?
    \item To which extent (if at all) can changes in probing accuracy be attributed to a change in linguistic knowledge encoded by the model?
\end{enumerate}

\noindent To answer these questions, we investigate three different pre-trained encoder models, BERT, RoBERTa, and ALBERT. We fine-tune them on sentence-level classification tasks from the GLUE benchmark \cite{wang2018glue} and evaluate the linguistic knowledge they encode leveraging three sentence-level probing tasks from the SentEval probing suite \cite{conneau-etal-2018-cram}. We focus on sentence-level probing tasks to measure linguistic knowledge encoded by a model for two reasons: 1) during fine-tuning we explicitly train a model to represent sentence-level context in its representations and 2) we are interested in the extent to which this affects existing sentence-level linguistic knowledge already present in a pre-trained model.

We find that while, indeed, fine-tuning affects a model's sentence-level probing accuracy and these effects are typically larger for higher layers, changes in probing accuracy vary depending on the encoder model, fine-tuning and probing task combination. Our results also show that sentence-level probing accuracy is highly dependent on the pooling method being used. \textbf{Only in very few cases, fine-tuning has a positive effect on probing accuracy that is larger than just using the pre-trained model with a strong pooling method.} Our findings suggest that changes in probing performance can not exclusively be attributed to an improved or deteriorated encoding of linguistic knowledge and should be carefully interpreted.  We present further evidence for this interpretation by investigating changes in the attention distribution and language modeling capabilities of fine-tuned models which constitute alternative explanations for changes in probing accuracy.

\section{Related Work}
\label{sec:related}

\paragraph{Probing} 

A large body of previous work focuses on analyses of the internal representations of neural models and the linguistic knowledge they encode \cite{shi-etal-2016-string, ettinger-etal-2016-probing, adi2016fine, belinkov-etal-2017-neural, hupkes2018visualisation}. In a similar spirit to these first works on probing, \citet{conneau-etal-2018-cram} were the first to compare different sentence embedding methods for the linguistic knowledge they encode. \citet{krasnowska-kieras-wroblewska-2019-empirical} extended this approach to study sentence-level probing tasks on English and Polish sentences. 

Alongside sentence-level probing, many recent works \cite{peters-etal-2018-deep, liu-etal-2019-linguistic, tenney2018what, lin-etal-2019-open, hewitt-manning-2019-structural} have focused on token-level probing tasks investigating more recent contextualized embedding models such as ELMo \cite{peters-etal-2018-deep}, GPT \cite{radford2019language}, and BERT \cite{devlin-etal-2019-bert}. Two of the most prominent works following this methodology are \citet{liu-etal-2019-linguistic} and \citet{tenney2018what}. While \citet{liu-etal-2019-linguistic} use linear probing classifiers as we do, \citet{tenney2018what} use more expressive, non-linear classifiers. However, in contrast to our work, most studies that investigate pre-trained contextualized embedding models focus on pre-trained models and not fine-tuned ones. Moreover, we aim to assess how probing performance changes with fine-tuning and how these changes differ based on the model architecture, as well as probing and fine-tuning task combination. 

\paragraph{Fine-tuning}

While fine-tuning pre-trained language models leads to a strong empirical performance across various supervised NLP downstream tasks \cite{wang2018glue}, fine-tuning itself \cite{dodge2020fine} and its effects on the representations learned by a pre-trained model are poorly understood. As an example,  \citet{phang2018sentence} show that downstream accuracy can benefit from an intermediate fine-tuning task, but leave the investigation of why certain tasks benefit from intermediate task training to future work. Recently, \citet{pruksachatkun2020intermediate} extended this approach using eleven diverse intermediate fine-tuning tasks. They view probing task performance after fine-tuning as an indicator of the acquisition of a particular language skill during intermediate task fine-tuning. This is similar to our work in the sense that probing accuracy is used to understand how fine-tuning affects a pre-trained model. \citet{talmor2019olmpics} try to understand whether the performance on downstream tasks should be attributed to the pre-trained representations or rather the fine-tuning process itself. They fine-tune BERT and RoBERTa on a large set of symbolic reasoning tasks and find that while RoBERTa generally outperforms BERT in its reasoning abilities, the performance of both models is highly context dependent. 

Most similar to our work is the contemporaneous work by \citet{merchant2020happens}. They investigate how fine-tuning leads to changes in the representations of a pre-trained model. In contrast to our work, their focus, however, lies on edge-probing \cite{tenney2018what} and structural probing tasks \cite{hewitt-manning-2019-structural} and they study only a single pre-trained encoder: BERT. We consider our work complementary to them since we study sentence-level probing tasks, use different analysis methods and investigate the impact of fine-tuning on three different pre-trained encoders: BERT, RoBERTa, and ALBERT.

\section{Methodology and Setup}
\label{sec:methodology-and-setup}

The focus of our work is on studying how fine-tuning affects the representations learned by a pre-trained model. We assess this change through sentence-level probing tasks. We focus on sentence-level probing tasks since during fine-tuning we explicitly train a model to represent sentence-level context in the CLS token. 

The fine-tuning and probing tasks we study concern different linguistic levels, requiring a model to focus more on syntactic, semantic or discourse information. The extent to which knowledge of a particular linguistic level is needed to perform well differs from task to task. For instance, to judge if the syntactic structure of a sentence is intact, no deep discourse understanding is needed. Our hypothesis is that if a pre-trained model encodes certain linguistic knowledge, this acquired knowledge should lead to a good performance on a probing task testing for the same linguistic phenomenon.  Extending this hypothesis to fine-tuning, one might argue that if fine-tuning introduces new or removes existing linguistic knowledge into/from a model, this should be reflected by an increase or decrease in probing performance.\footnote{\citet{merchant2020happens} follow a similar reasoning. They find that fine-tuning on dependency parsing task leads to an improvement on the constituents probing task and attribute this to the improved linguistic knowledge. Similarly, \citet{pruksachatkun2020intermediate} view probing task performance as ``an indicator for the acquisition of a particular language skill."} However, we argue that \textbf{encoding or forgetting linguistic knowledge is not necessarily the only explanation for observed changes in probing accuracy}. Hence, the goal of our work is to test the above-stated hypotheses assessing the interaction between fine-tuning and probing tasks across three different encoder models.

\subsection{Fine-tuning tasks}

\begin{table}[t]
    \centering
    \resizebox{.98\columnwidth}{!}{%
    \begin{tabular}{lcccc} 
        \toprule
        \multirow{2}{*}{\textbf{Model}} & \multicolumn{4}{c}{\textbf{Task}}\\ \cmidrule{2-5}
         & CoLA & SST-2  & RTE & SQuAD   \\
        \midrule 
        \citet{devlin-etal-2019-bert} & $52.1$ & $93.5$  & $66.4$ & $80.8$/$88.5$ \\
        \midrule 
        BERT & $59.5$ & $92.4$ & $64.6 $ & $78.6$/$86.5$  \\
        RoBERTa & $60.3$ & $93.6$ & $73.6$  & $81.7$/$89.3$  \\ 
        ALBERT & $45.8$ & $88.5$  & $69.6$  & $79.9$/$87.6$  \\
        
        \bottomrule
    \end{tabular}
    }
    \caption{Fine-tuning performance on the development set on selected down-stream tasks. For comparison we also report the fine-tuning accuracy of BERT-base-cased as reported by \citet{devlin-etal-2019-bert} on the test set of each of the tasks taken from the GLUE and SQuAD leaderboards. We report Matthews correlation coefficient for CoLA, accuracy for SST-2 and RTE, and exact match (EM) and $F_1$ score for SQuAD.
    }
    \label{tab:fine-tuning-results}
\end{table}

We study three fine-tuning tasks taken from the GLUE benchmark \cite{wang2018glue}. All the tasks are sentence-level classification tasks and cover different levels of linguistic phenomena. Additionally, we study models fine-tuned on SQuAD \cite{rajpurkar-etal-2016-squad} a widely used question answering dataset. Statistics for each of the tasks can be found in the Appendix.

\paragraph{CoLA} The Corpus of Linguistic Acceptability \cite{warstadt2018neural} is an acceptability task which tests a model's knowledge of grammatical concepts. We expect that fine-tuning on CoLA results in changes in accuracy on a syntactic probing task.\footnote{CoLA contains sentences with syntactic, morphological and semantic violations.
However, only about 15\% of the sentences are labeled with morphological and semantic violations. 
Hence, we suppose that fine-tuning on CoLA should increase a model's sensitivity to syntactic violations to a greater extent.}

\paragraph{SST-2} The Stanford Sentiment Treebank \cite{socher-etal-2013-recursive}. We use the binary version  where the task is to categorize movie reviews to have either positive or negative valence. Making sentiment judgments requires knowing the meanings of isolated words and combining them on the sentence and discourse level (e.g. in case of irony). Hence, we expect to see a difference for semantic and/or discourse probing tasks when fine-tuning on SST-2.

\paragraph{RTE} The Recognizing Textual Entailment dataset is a collection of sentence-pairs in either neutral or entailment relationship collected from a series of annual textual entailment challenges \cite{10.1007/11736790_9, barhaim-rte, 10.5555/1654536.1654538, Bentivogli09thefifth}.
The task requires a deeper understanding of the relationship of two sentences, hence, fine-tuning on RTE might affect the accuracy on a discourse-level probing task.

\paragraph{SQuAD} The Stanford Questions Answering Dataset \cite{rajpurkar-etal-2016-squad} is a popular extractive reading comprehension dataset. The task involves a broader discourse understanding as a model trained on SQuAD is required to extract the answer to a question from an accompanying paragraph.

\subsection{Probing Tasks}

We select three sentence-level probing tasks from the SentEval probing suit \cite{conneau-etal-2018-cram}, testing for syntactic, semantic and broader discourse information on the sentence-level.

\paragraph{bigram-shift} is a syntactic binary classification task that tests a model's sensitivity to word order. The dataset consists of intact and corrupted sentences, where for corrupted sentences, two random adjacent words have been inverted.

\paragraph{semantic-odd-man-out} tests a model's sensitivity to semantic incongruity on a collection of sentences where random verbs or nouns are replaced by another verb or noun.

\paragraph{coordination-inversion} is a collection of sentences made out of two coordinate clauses. In half of the sentences, the order of the clauses is inverted. Coordinate-inversion tests for a model's broader discourse understanding.

\begin{figure*}[ht!]
    \centering
    \begin{subfigure}[t]{.32\textwidth}
        \centering
        \includegraphics[width=1.0\textwidth]{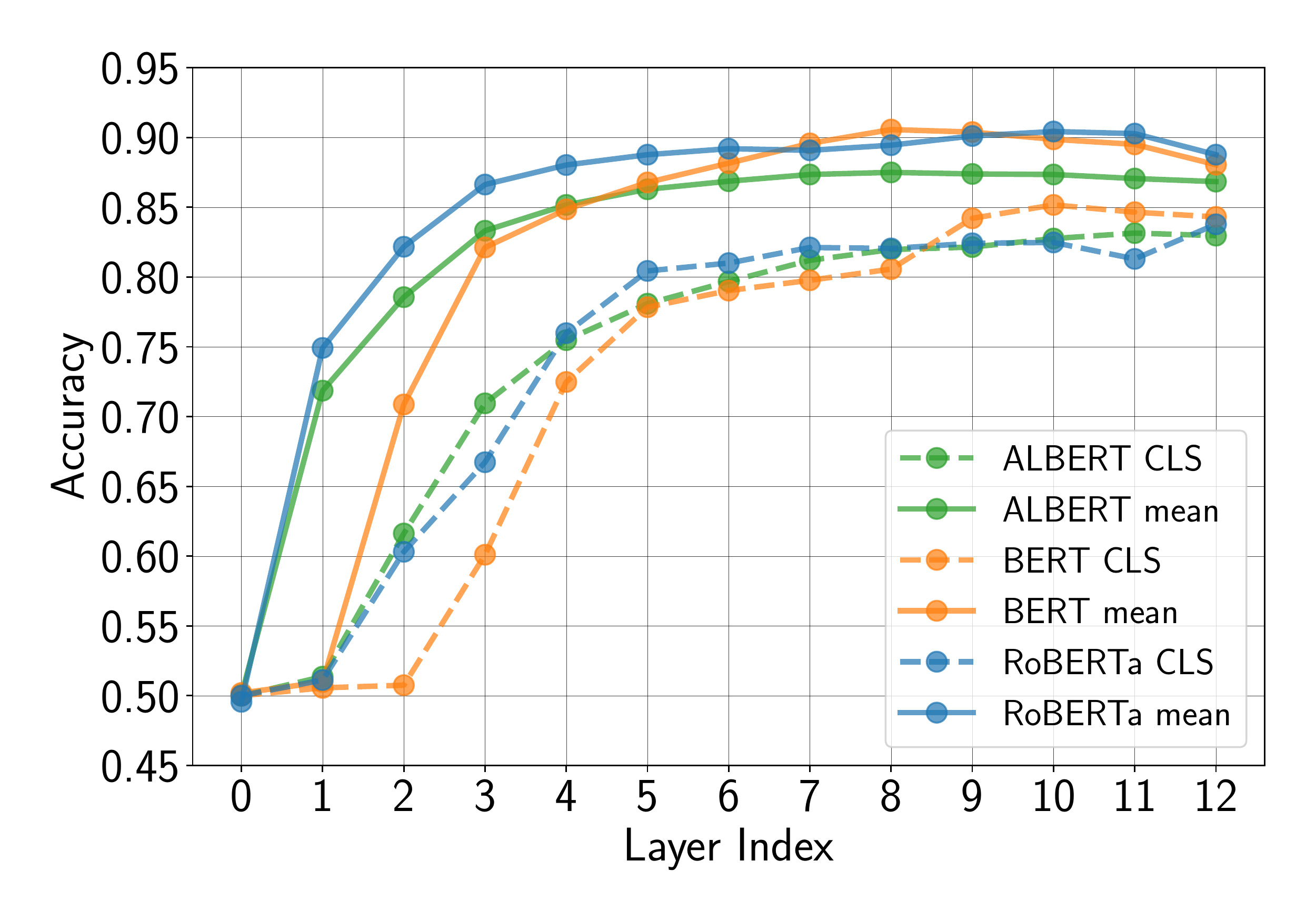}
        \caption{bigram-shift}
    \end{subfigure}%
    ~
    \begin{subfigure}[t]{.32\textwidth}
        \centering
        \includegraphics[width=1.0\textwidth]{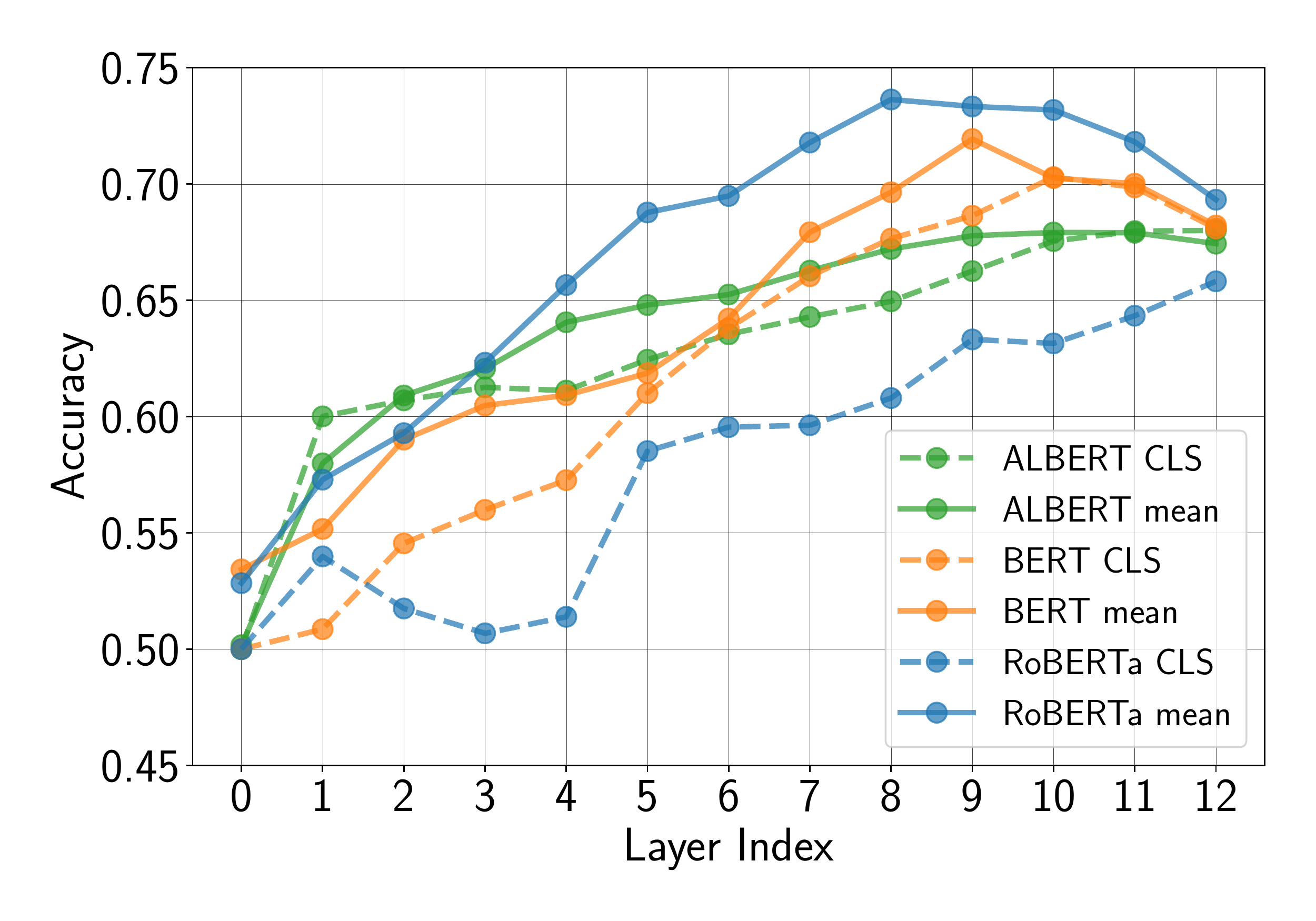}
        \caption{coordination-inversion}
    \end{subfigure}%
    ~
    \begin{subfigure}[t]{.32\textwidth}
        \centering
        \includegraphics[width=1.0\textwidth]{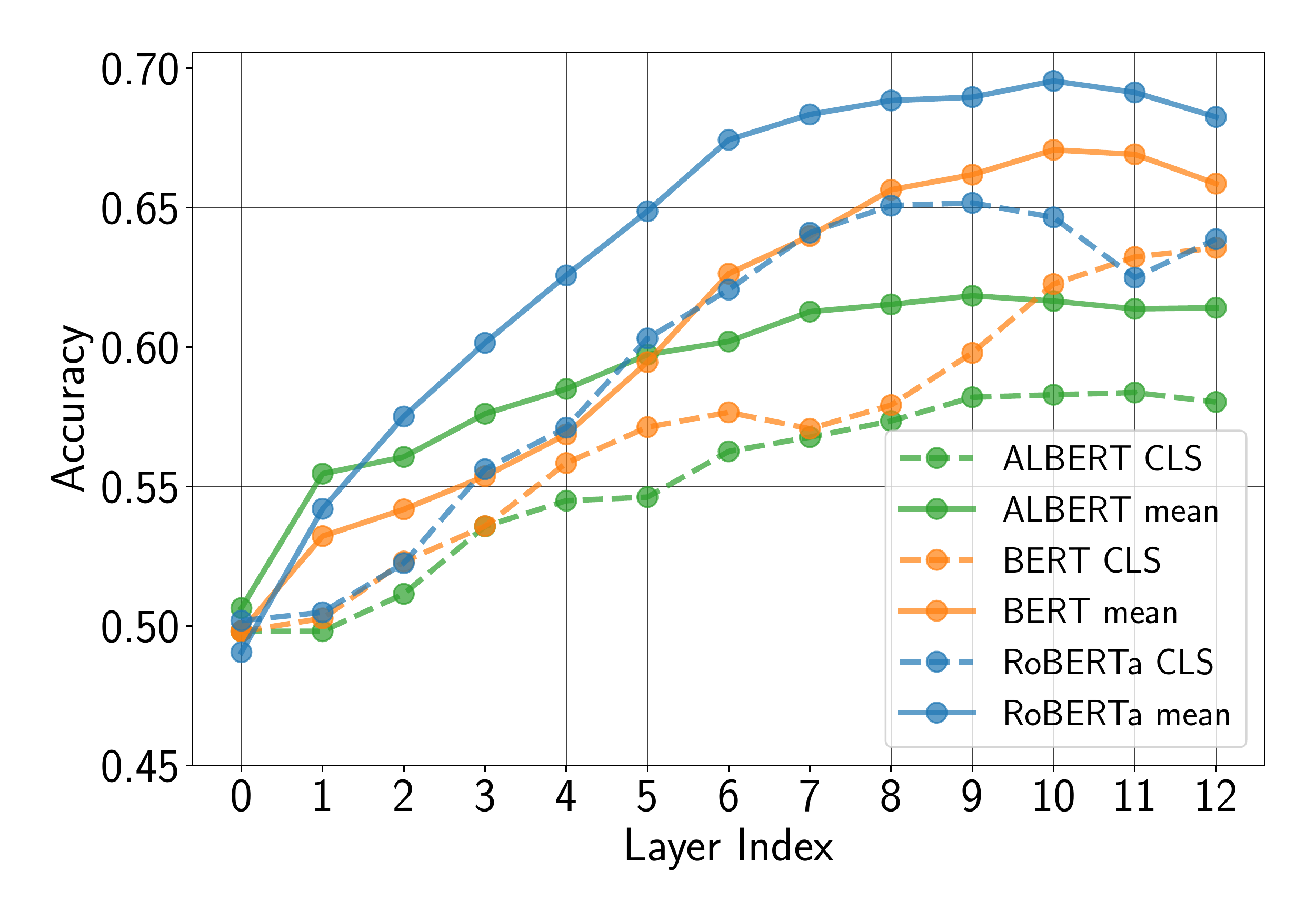}
        \caption{odd-man-out}
    \end{subfigure}%

    \caption{Layer-wise probing accuracy on bigram-shift, coordination inversion, and odd-man-out for BERT, RoBERTa, and ALBERT. For all models mean-pooling (solid lines) consistently improves probing accuracy compared to CLS-pooling (dashed-lines) highlighting the importance of sentence-level information for each of the tasks.
    }
    \label{fig:pre-trained-accs}
\end{figure*}

\subsection{Pre-trained Models}

It is unclear to which extent findings on the encoding of certain linguistic phenomena generalize from one pre-trained model to another. Hence, we examine three different pre-trained encoder models in our experiments.

\paragraph{BERT} \cite{devlin-etal-2019-bert} is a transformer-based model \cite{NIPS2017_7181} jointly trained on masked language modeling and next-sentence-prediction -- a sentence-level binary classification task. BERT was trained on the Toronto Books corpus and the English portion of Wikipedia.  
We focus on the \textbf{BERT-base-cased} model which consists of 12 hidden layers and will refer to it as BERT in the following. 

\paragraph{RoBERTa} \cite{liu2019roberta} is a follow-up version of BERT which differs from BERT in a few crucial aspects, including using larger amounts of training data and longer training time. The aspect  that is most relevant in the context of this work is that RoBERTa was pre-trained without a sentence-level objective, minimizing only the masked language modeling objective. As with BERT we will consider the base model, \textbf{RoBERTa-base}, for this study and refer to it as RoBERTa.

\paragraph{ALBERT} \cite{Lan2020ALBERT:} is another recently proposed transformer-based pre-trained masked language model. In contrast to both BERT and RoBERTa, it makes heavy use of parameter sharing. That is, ALBERT ties the weight matrices across all hidden layers effectively applying the same non-linear transformation on every hidden layer. Additionally, similar to BERT, ALBERT uses a sentence-level pre-training task. We will use the base model \textbf{ALBERT-base-v1} and refer to it as ALBERT throughout this work. 

\subsection{Fine-tuning and Probing Setup}
\label{sec:fine-tuning-and-probing-setup}

\paragraph{Fine-tuning}

For fine-tuning, we follow the default setup proposed by \citet{devlin-etal-2019-bert}. A single randomly initialized task-specific classification layer is added on top of the pre-trained encoder. As input, the classification layer receives $\mathbf{z} = \tanh{(\mathbf{W}\mathbf{h} + \mathbf{b})}$, where $\mathbf{h}$ is the hidden representation of the first token on the last hidden layer and $\mathbf{W}$ and $\mathbf{b}$ are the randomly initialized parameters of the classifier.\footnote{For BERT and ALBERT $\mathbf{h}$ corresponds to the hidden state of the [CLS] token. For RoBERTa the first token of every sentence is the $<$s$>$ token. We will refer to both of them as CLS token.} During fine-tuning all model parameters are updated jointly. We train for 3 epochs on CoLA and for 1 epoch on SST-2, using a learning rate of $\num{2e-05}$. The learning rate is linearly increased for the first $10\%$ of steps (warmup) and kept constant afterwards. An overview of all hyper-parameters for each model and task can be found in the Appendix. Fine-tuning performance on the development set of each of the tasks can be found in Table \ref{tab:fine-tuning-results}.

\paragraph{Probing}

\begin{table*}[ht!]
    \centering
        \begin{subfigure}[t]{\textwidth}
            \centering
                \begin{tabular}{lrrrrrrrr} 
                    \toprule
                    \multirow{4}{*}{\textbf{Probing Task}} &  \multicolumn{8}{c}{\textbf{\textcolor{bert-orange}{BERT-base-cased}}}\\ \cmidrule{2-9}
                    & \multicolumn{4}{c}{CLS-pooling} & \multicolumn{4}{c}{mean-pooling}  \\ \cmidrule{2-9}
                    & \multicolumn{2}{c}{CoLA} & \multicolumn{2}{c}{SST-2} & \multicolumn{2}{c}{CoLA} & \multicolumn{2}{c}{SST-2}  \\ 
                     &  0 -- 6 & 7 -- 12 &  0 -- 6 & 7 -- 12 & 0 -- 6 & 7 -- 12 & 0 -- 6 & 7 -- 12 \\
                    \midrule
                    bigram-shift &  $0.07$ & $4.73$ & $-1.02$ & $-4.63$ & $0.23$ & $1.45$ & $-0.37$ & $-3.24$  \\
                    coordinate-inversion &  $-0.10$ & $1.90$ & $-0.25$ & $-1.15$ & $0.14$ & $0.29$ & $-0.48$ & $-0.85$  \\
                    odd-man-out &  $-0.20$ & $0.26$ & $-0.02$ & $-1.28$ & $-0.34$ & $-0.29$ & $-0.30$ & $-1.09$ \\
                \end{tabular}
        \end{subfigure}\vspace{.5em}%
        \\
        \begin{subfigure}[t]{\textwidth}
            \centering
                \begin{tabular}{lrrrrrrrr} 
                        \toprule
                        \multirow{4}{*}{\textbf{Probing Task}} &  \multicolumn{8}{c}{\textbf{\textcolor{roberta-blue}{RoBERTa-base}}}\\ \cmidrule{2-9}
                        & \multicolumn{4}{c}{CLS-pooling} & \multicolumn{4}{c}{mean-pooling}  \\ \cmidrule{2-9}
                    & \multicolumn{2}{c}{CoLA} & \multicolumn{2}{c}{SST-2} & \multicolumn{2}{c}{CoLA} & \multicolumn{2}{c}{SST-2}  \\ 
                         &  0 -- 6 & 7 -- 12 &  0 -- 6 & 7 -- 12 & 0 -- 6 & 7 -- 12 & 0 -- 6 & 7 -- 12  \\
                        \midrule
                        bigram-shift &  $0.58$ & $5.35$ & $-2.41$ & $-7.22$ & $0.69$ & $1.74$ & $-0.23$ & $-4.87$ \\
                        coordinate-inversion &  $-0.72$ & $1.84$ & $-1.28$ & $-0.63$ & $-0.22$ & $0.02$ & $-0.18$ & $-3.83$ \\
                        odd-man-out &  $-0.66$ & $1.05$ & $-1.09$ & $-2.40$ & $-0.08$ & $-0.55$ & $-0.46$ & $-3.61$ \\
                    \end{tabular}
        \end{subfigure}\vspace{.5em}%
        \\
        \begin{subfigure}[t]{\textwidth}
            \centering
                \begin{tabular}{lrrrrrrrr} 
                    \toprule
                    \multirow{4}{*}{\textbf{Probing Task}} &  \multicolumn{8}{c}{\textbf{\textcolor{albert-green}{ALBERT-base-v1}}}\\ \cmidrule{2-9}
                    & \multicolumn{4}{c}{CLS-pooling} & \multicolumn{4}{c}{mean-pooling}  \\ \cmidrule{2-9}
                    & \multicolumn{2}{c}{CoLA} & \multicolumn{2}{c}{SST-2} & \multicolumn{2}{c}{CoLA} & \multicolumn{2}{c}{SST-2}  \\ 
                     &  0 -- 6 & 7 -- 12 &  0 -- 6 & 7 -- 12 & 0 -- 6 & 7 -- 12 & 0 -- 6 & 7 -- 12  \\
                    \midrule
                    bigram-shift &  $1.55$ & $3.39$ & $-1.94$ & $-5.15$ & $0.26$ & $0.66$ & $-0.70$ & $-2.73$ \\
                    coordinate-inversion &  $-0.69$ & $-1.53$ & $-1.07$ & $\textit{-2.87}$ & $-0.07$ & $-1.19$ & $-0.35$ & $-1.53$  \\
                    odd-man-out &  $-0.42$ & $-1.39$ & $-0.90$ & $-2.75$ & $-0.27$ & $-1.40$ & $-0.60$ & $-2.82$  \\
                    \bottomrule
                \end{tabular}
        \end{subfigure}%
    \caption{Change in probing accuracy $\Delta$ (in $\%$) of \textbf{CoLA} and \textbf{SST-2} fine-tuned models compared to the pre-trained models when using CLS and mean-pooling. We average the difference in probing accuracy over two different layers groups: layers 0 to 6 and layers 7 to 12.}
    \label{tab:aggregated-acc-deltas-cls}
\end{table*}

For probing, our setup largely follows that of previous works \cite{tenney2018what, liu-etal-2019-linguistic, hewitt-liang-2019-designing} where a \textit{probing classifier} is trained on top of the contextualized embeddings extracted from a pre-trained or -- as in our case -- fine-tuned encoder model. Notably, we train  \textit{linear} (logistic regression) probing classifiers and use two different \textit{pooling methods} to obtain sentence embeddings from the encoder hidden states: \textbf{CLS-pooling}, which simply returns the hidden state corresponding to the first token of the sentence and \textbf{mean-pooling} which computes a sentence embedding as the mean over all hidden states. We do this to assess the extent to which the CLS token captures sentence-level context. We use linear probing classifiers because intuitively we expect that if a linguistic feature is useful for a fine-tuning task, it should be linearly separable in the embeddings. For all probing tasks, we measure layer-wise accuracy to investigate how the linear separability of a particular linguistic phenomenon changes across the model. In total, we train 390 probing classifiers on top of 12 pre-trained and fine-tuned encoder models.

\paragraph{Implementation}

Our experiments are implemented in PyTorch \cite{NIPS2019_9015} and we use the pre-trained models provided by the HuggingFace transformers library \cite{Wolf2019HuggingFacesTS}. Code to reproduce our results and figures is available online: \url{https://github.com/uds-lsv/probing-and-finetuning}

\section{Experiments}
\label{sec:experiments}

\subsection{Probing Accuracy}
\label{sec:probing-accuracy}

Figure \ref{fig:pre-trained-accs} shows the layer-wise probing accuracy of BERT, RoBERTa, and ALBERT on each of the probing tasks. These results establish baselines for our comparison with fine-tuned models below. Consistent with previous work \cite{krasnowska-kieras-wroblewska-2019-empirical}, we observe that mean-pooling generally outperforms CLS-pooling across all probing tasks, highlighting the importance of sentence-level context for each of the probing tasks. We also find that for \textit{bigram-shift} probing accuracy is substantially larger than that for coordination-inversion and odd-man-out. Again, this is consistent with findings in previous works \cite{tenney2018what, liu-etal-2019-linguistic, tenney-etal-2019-bert} reporting better performance on syntactic than semantic probing tasks. 

When comparing the three encoder models, we observe some noticeable differences. On \textit{odd-man-out}, ALBERT performs significantly worse than both BERT and RoBERTa, with RoBERTa performing best across all layers. We attribute the poor performance of ALBERT to the fact that it makes heavy use of weight-sharing, effectively applying the same non-linear transformation on all layers. We also observe that on \textit{coordination-inversion}, RoBERTa with CLS pooling performs much worse than both BERT and ALBERT with CLS pooling. We attribute this to the fact that RoBERTa lacks a sentence-level pre-training objective and the CLS token hence fails to capture relevant sentence-level information for this particular probing task. The small differences in probing accuracy for BERT and ALBERT when comparing CLS to mean-pooling and the fact that RoBERTa with mean-pooling outperforms all other models on \textit{coordination-inversion} is providing evidence for this interpretation.

\subsection{How does Fine-tuning affect Probing Accuracy?}
\label{sec:probing-and-fine-tuning}

Having established baselines for the probing accuracy of the pre-trained models, we now turn to the question of how it is affected by fine-tuning. Table \ref{tab:aggregated-acc-deltas-cls} shows the effect of fine-tuning on CoLA and SST-2 on the layer-wise accuracy for all three encoder models across the three probing tasks.  Results for RTE and SQuAD can be found in Table \ref{tab:aggregated-acc-deltas-cls-mean-rte-squad11} in the Appendix. \textbf{For all models and tasks we find that fine-tuning has mostly an effect on higher layers, both positive and negative.}
The impact varies depending on the fine-tuning/probing task combination and underlying encoder model.

\paragraph{Positive Changes in Accuracy:} 
Fine-tuning on CoLA results in a substantial improvement on the \textit{bigram-shift} probing task for all the encoder models; fine-tuning on RTE improves the \textit{coordination-inversion} accuracy for RoBERTa. This finding is in line with our expectations: \textit{bigram-shift} and CoLA require syntactic level information, whereas \textit{coordination-inversion} and RTE require a deeper discourse-level understanding.
However, when taking a more detailed look, this reasoning becomes questionable:
The improvement is only visible when using  CLS-pooling and becomes negligible when probing with mean-pooling. Moreover, the gains are not large enough to improve significantly over the mean-pooling baseline (as shown by the stars and the second y-axis in Figure \ref{fig:aggregated-acc-deltas-cola-and-stars}). This suggests that adding new linguistic knowledge is not necessarily the \textit{only} driving force behind the improved probing accuracy and we provide evidence for this reasoning in Section \ref{sec:attention}.

\paragraph{Negative Changes in Accuracy:}
Across all models and pooling methods, fine-tuning on SST-2 has a negative impact on probing accuracy on \textit{bigram-shift} and \textit{odd-man-out}, and the decrease in probing accuracy is particularly large for RoBERTa.
Fine-tuning on SQuAD follows a similar trend: it has a negative effect on probing accuracy on \textit{bigram-shift} and \textit{odd-man-out} for both CLS- and mean-pooling (see Table \ref{tab:aggregated-acc-deltas-cls-mean-rte-squad11}), while the impact on \textit{coordination-inversion} is negligible.
We argue that this strong negative impact on probing accuracy is the consequence of more dramatic changes in the representations.
We investigate this issue further in Section~\ref{sec:perplexity}. \\

\noindent Changes in probing accuracy for other fine-tuning/probing combinations are not substantial, which suggests that representations did not change significantly with regard to the probed information.

\begin{figure*}[t!]
    \centering
    \begin{subfigure}[t]{.40\textwidth}
        \includegraphics[width=\linewidth]{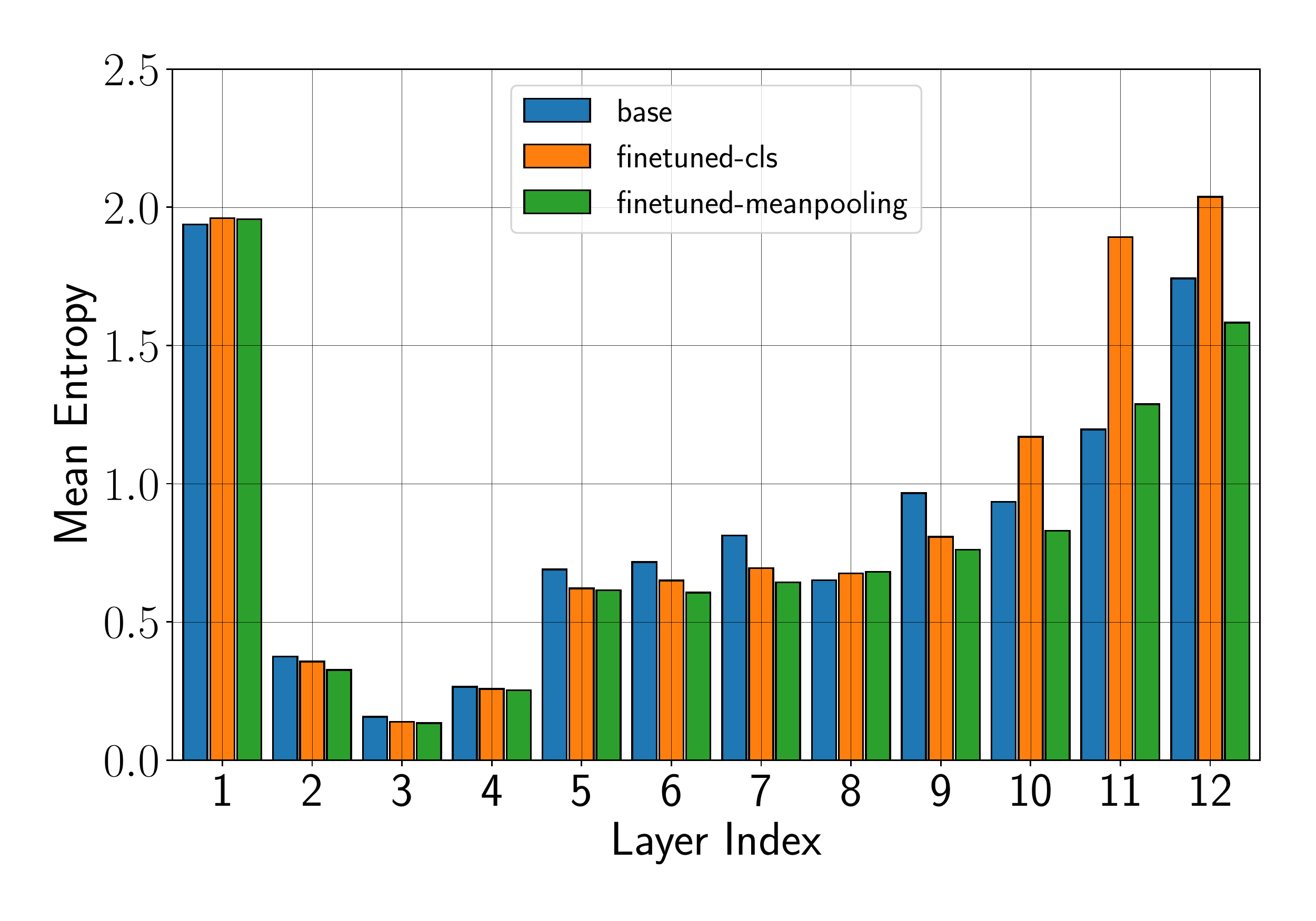}
        \caption{Entropy}
        \label{fig:attentions_entropy}
    \end{subfigure}
    ~
    \begin{subfigure}[t]{.40\textwidth}
        \includegraphics[width=\linewidth]{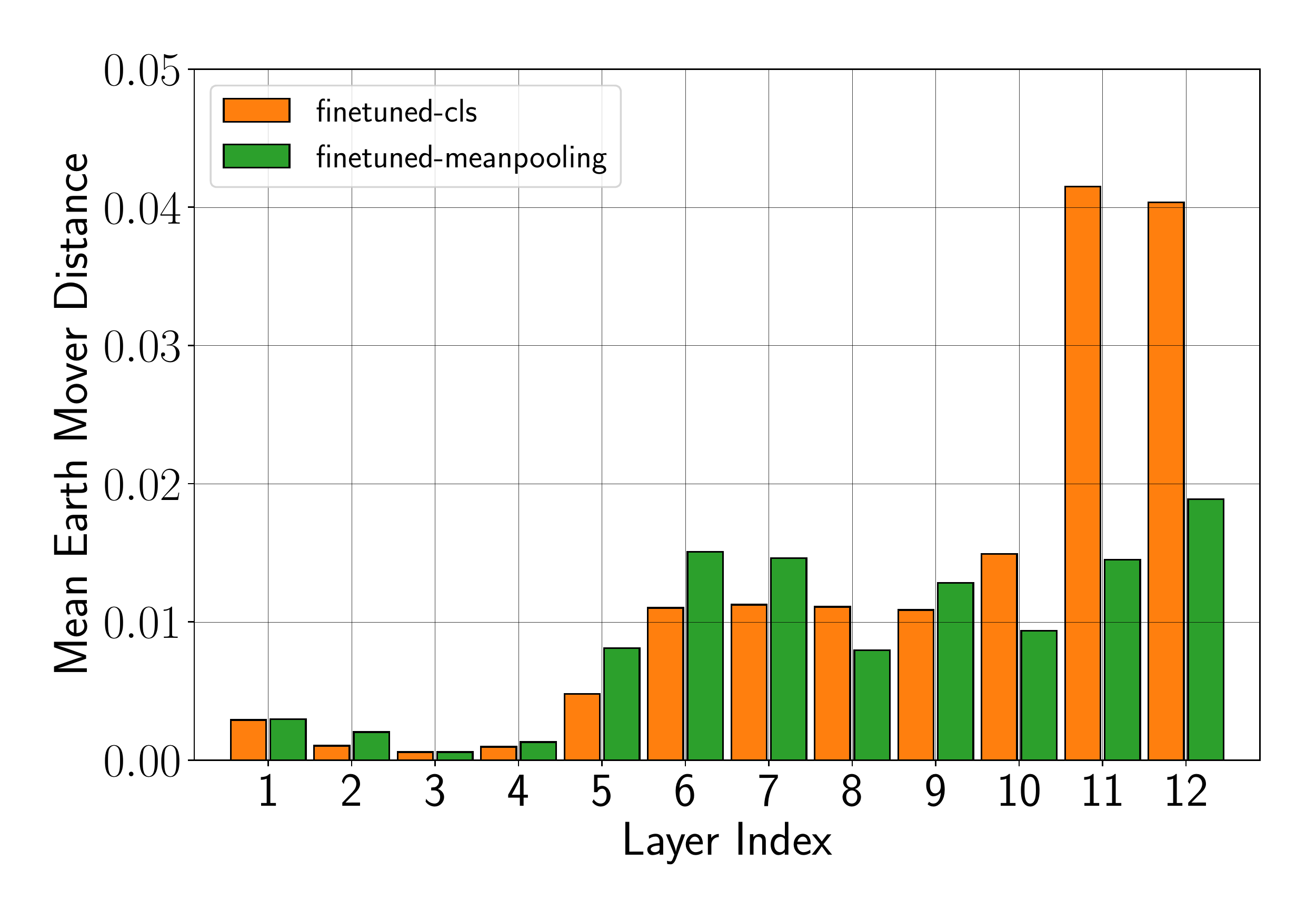}
        \caption{Earth-Mover Distance}
        \label{fig:attentions_earthmover}
    \end{subfigure}
    \caption{Entropy and Earth mover's distance of the attention for the CLS token for each layer with the RoBERTa model on the bigram-shift dataset. The mean over all input sequences and the mean over all attention heads of a layer are taken. The Earth Mover Distance is computed between the base model and each fine-tuned model.}
\end{figure*}

\section{What Happens During Fine-tuning?}
\label{sec:analysis}

In the previous part, we saw the effects of different fine-tuning approaches on model performance. This opens the question for their causes. In this section, we study two hypotheses that go towards explaining these effects.

\subsection{Analyzing Attention Distributions}
\label{sec:attention}

If the improvement in probing accuracy with CLS-pooling can be attributed to a better sentence representation in the CLS token, this can be due to a corresponding change in a model's attention distribution. The model might change the attention of the CLS token to cover more tokens and with this build a better representation of the whole sentence.

To study this hypothesis, we fine-tune RoBERTa on CoLA using two different methods: the default CLS-pooling approach and mean-pooling (cf. Section \ref{sec:fine-tuning-and-probing-setup}). We compare the layer-wise attention distribution on bigram-shift after fine-tuning to that data. We expect to see more profound changes for CLS-pooling than for mean-pooling. To investigate how the attention distribution changes, we analyze its entropy, i.e. 

\begin{equation}
   H_j = \sum_i a_j(x_i) \cdot log\left( a_j(x_i) \right)
\end{equation}

\noindent where $x_i$ is the $i$-th token of an input sequence and $a(x_i)$ the corresponding attention at position $j$ given to it by a specific attention head. Entropy is maximal when the attention is uniform over the whole input sequence and minimal if the attention head focuses on just one input token. 

Figure \ref{fig:attentions_entropy} shows the mean entropy for the CLS token (i.e. $H_0$) before and after fine-tuning. We observe a large increase in entropy in the last three layers when fine-tuning on the CLS token (orange bars). This is consistent with our interpretation that, during fine-tuning, the CLS token learns to take more sentence-level information into account, therefore being required to spread its attention over more tokens. For mean-pooling (green bars) this might not be required as taking the mean over all token-states could already provide sufficient sentence-level information during fine-tuning. Accordingly, there are only small changes in the entropy for mean-pooling, with the mean entropy actually decreasing in the last layer.

Entropy alone is, however, not sufficient to analyze changes in the attention distribution. Even when the amount of entropy is similar, the underlying attention distribution might have changed. Figure \ref{fig:attentions_earthmover}, therefore, compares the attentions of an attention head for an input sequence before and after fine-tuning using \textit{Earth mover's distance} \cite{710701}. We find that, similarly to the entropy results, changes in attention tend to increase with the layer number and again, the largest change of the attention distribution is visible for the first token for layer 11 and 12 when pooling on the CLS-token, while the change is much smaller for mean-pooling. This affirms our hypothesis that improvements in the fine-tuning with CLS-pooling can be attributed to a change in the attention distribution which is less necessary for the mean-pooling.

\begin{figure*}[ht!]
    \centering
    \begin{subfigure}[t]{.40\textwidth}
        \centering
        \includegraphics[width=1.0\textwidth]{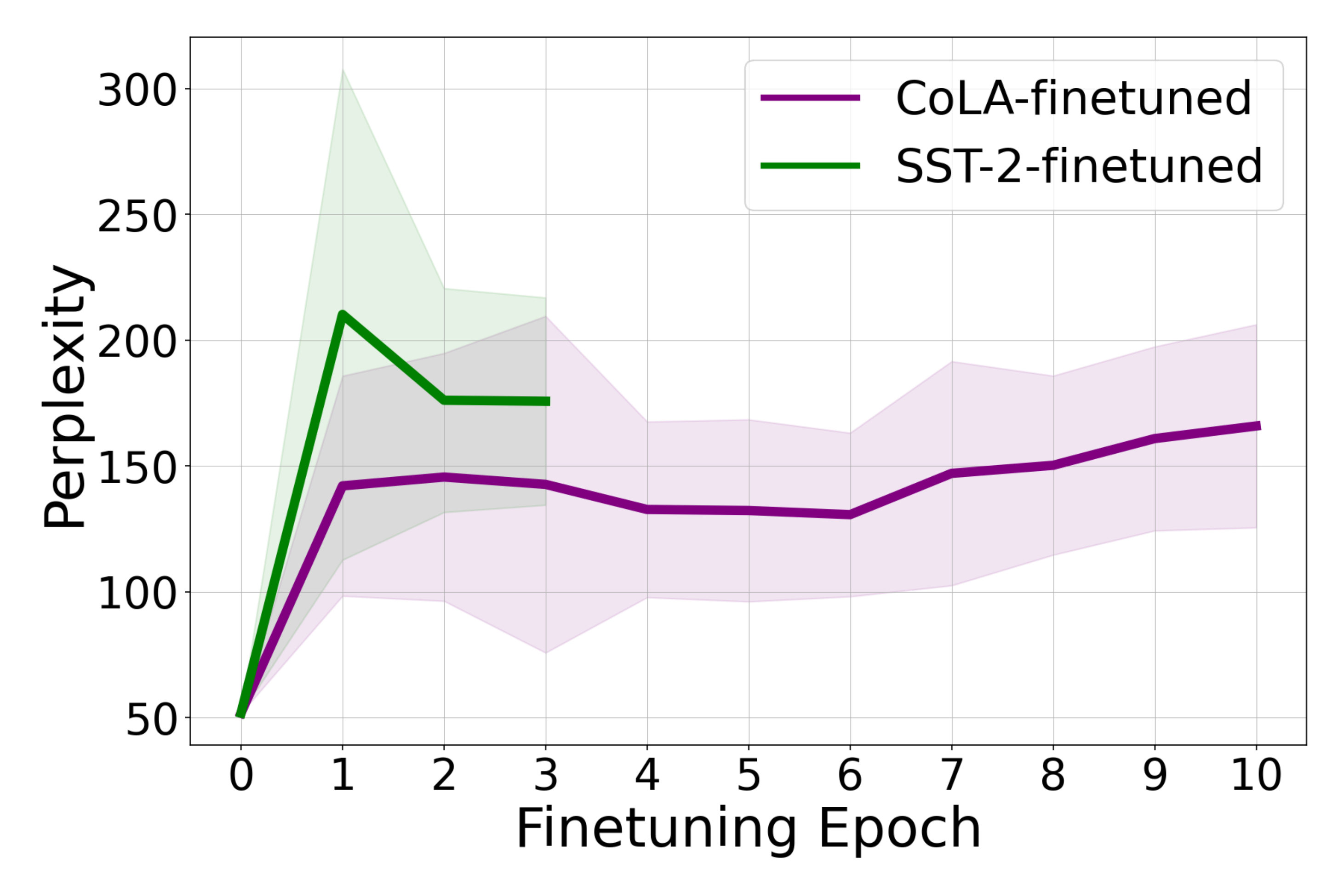}
        \caption{RoBERTa-base}
        \label{fig:roberta_checkpoints}
    \end{subfigure}%
    ~ 
    \begin{subfigure}[t]{.40\textwidth}
        \centering
        \includegraphics[width=1.0\textwidth]{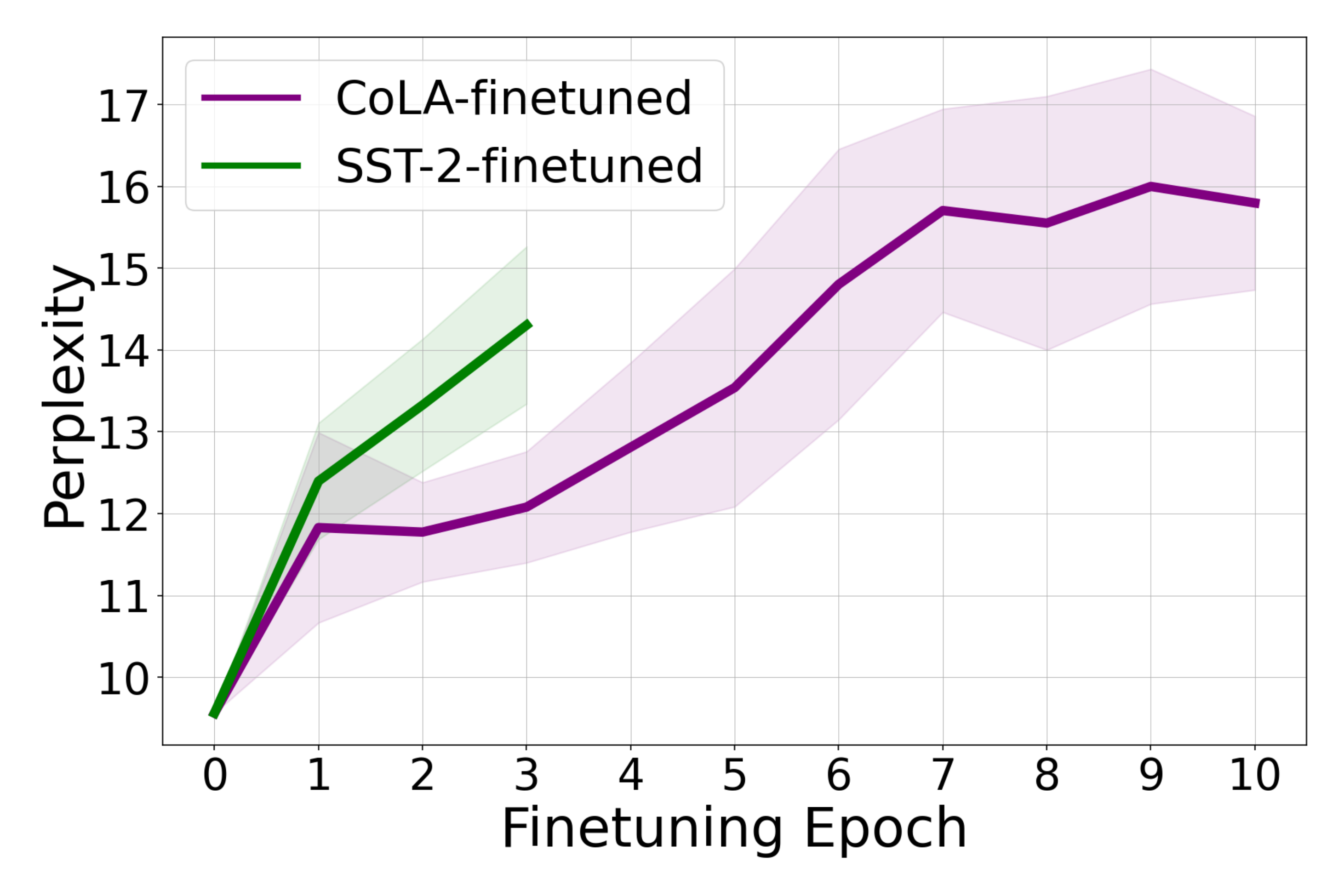}
        \caption{BERT-base-cased}
        \label{fig:bert_checkpoints}
    \end{subfigure}%
    \ 
    \begin{subfigure}[t]{.40\textwidth}
        \centering
        \includegraphics[width=1.0\textwidth]{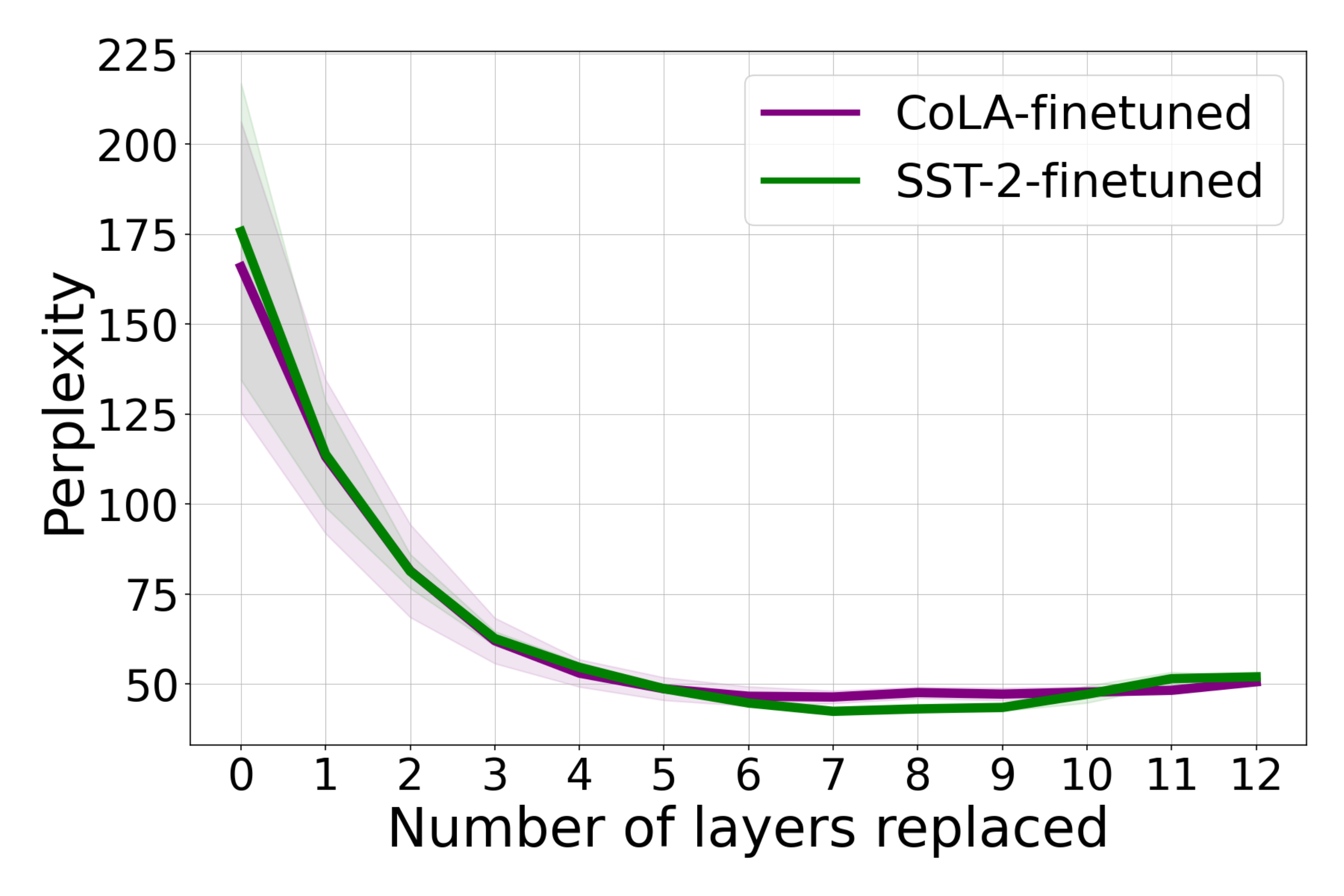}
        \caption{RoBERTa-base}
        \label{fig:roberta_layers}
    \end{subfigure}%
    ~
    \begin{subfigure}[t]{.40\textwidth}
        \centering
        \includegraphics[width=1.0\textwidth]{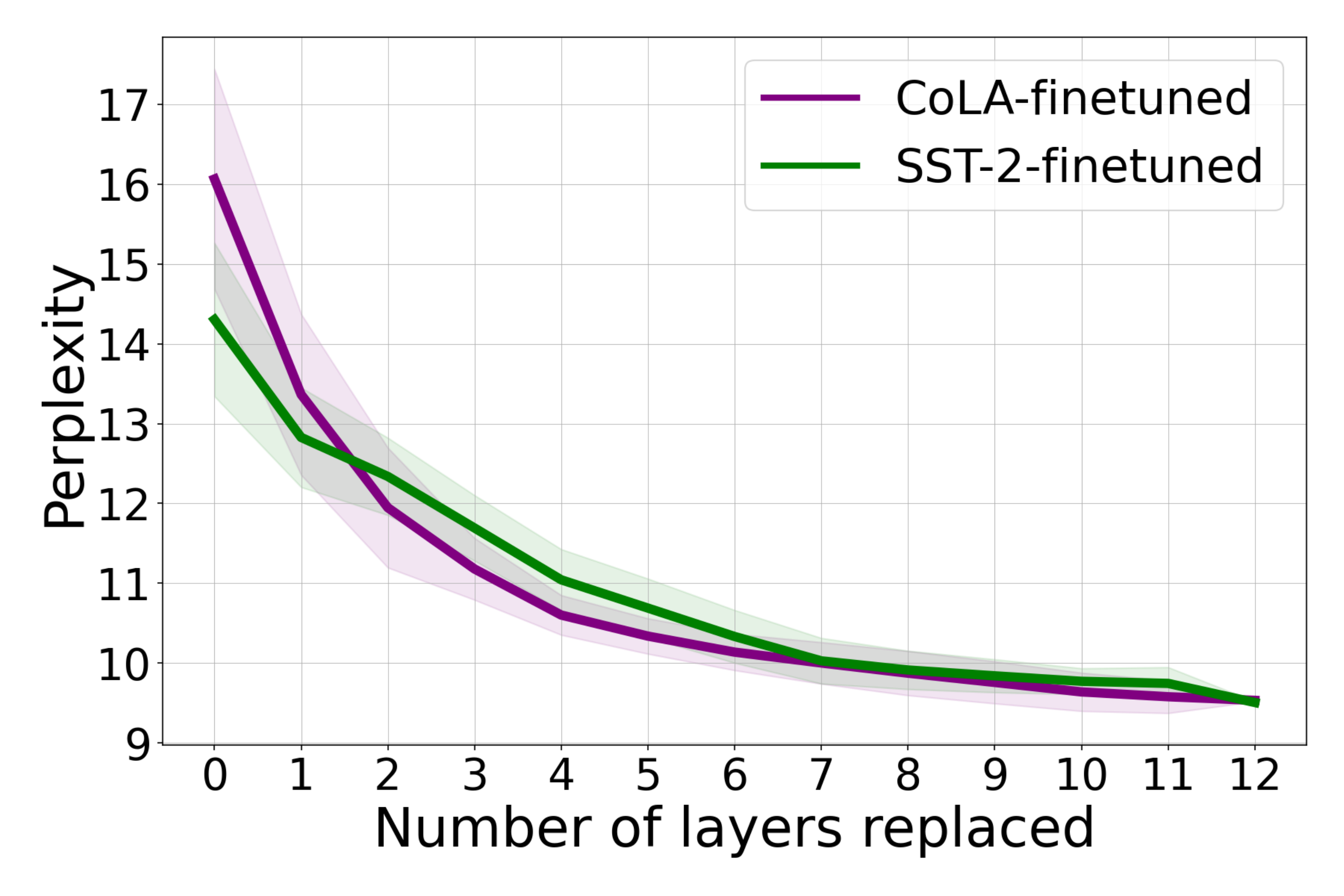}
        \caption{BERT-base-cased}
        \label{fig:bert_layers}
    \end{subfigure}%
    \caption{Perplexity on Wikitext-2 of models consisting of a fine-tuned encoder and a pre-trained MLM-head. Plots (a) and (b) show how perplexity changes over the course of fine-tuning with epoch 0 showing the perplexity of the pre-trained model. (c) and (d) show how perplexity changes when a number of last layers of the fine-tuned encoder are replaced with corresponding layers from the pre-trained model. Note the different y-axes for RoBERTa and BERT.}
    \label{fig:mlm-eval}
\end{figure*}

\subsection{Analyzing MLM Perplexity} 
\label{sec:perplexity}

If fine-tuning has more profound effects on the representations of a pre-trained model potentially introducing or removing linguistic knowledge, we expect to see larger changes to the language modeling abilities of the model when compared to the case where fine-tuning just changes the attention distribution of the CLS token. 

For this, we analyze how fine-tuning on CoLA and SST-2 affect the language modeling abilities of a pre-trained model. A change in perplexity should reveal if the representations of the model did change during fine-tuning and we expect this change to be larger for SST-2 fine-tuning where we observe a large negative increase in probing accuracy.

For the first experiment, we evaluate the pre-trained masked language model heads of BERT and RoBERTa on the Wikitext-2 test set \cite{Merity2017PointerSM} and compare it to the masked-language modeling perplexity, hereafter perplexity, of fine-tuned models.\footnote{Note that perplexity results are not directly comparable between BERT and RoBERTa since both models have different vocabularies. However, what we are interested in is rather how perplexity changes with fine-tuning.} In the second experiment, we test which layers contribute most to the change in perplexity and replace layers of the fine-tuned encoder by pre-trained layers, starting from the last layer. For both experiments, we evaluate the perplexity of the resulting model using the pre-trained masked language modeling head. We fine-tune and evaluate each model 5 times, and report the mean perplexity as well as standard deviation. Our reasoning is that if fine-tuning leads to dramatic changes to the hidden representations of a model, the effects should be reflected in the perplexity.

\paragraph{Perplexity During Fine-tuning}

Figure \ref{fig:roberta_checkpoints} and \ref{fig:bert_checkpoints} show how the perplexity of a pre-trained model changes during fine-tuning. Both BERT and RoBERTa show a similar trend where perplexity increases with fine-tuning. Interestingly, for RoBERTa the increase in perplexity after the first epoch is much larger compared to BERT. Additionally, our results show that for both models the increase in perplexity is larger when fine-tuning on SST-2. This confirms our hypothesis and also our findings from Section \ref{sec:experiments} suggesting that fine-tuning on SST-2 has indeed more dramatic effects on the representations of both models compared to fine-tuning on CoLA.

\paragraph{Perplexity When Replacing Fine-tuned Layers}

While fine-tuning leads to worse language modeling abilities for both CoLA and SST-2, it is not clear from the first experiment alone which layers are responsible for the increase in perplexity. Figure \ref{fig:roberta_layers} and \ref{fig:bert_layers} show the perplexity results when replacing fine-tuned layers with pre-trained ones starting from the last hidden layer. Consistent with our probing results in Section \ref{sec:experiments}, we find that the \textbf{changes that lead to an increase in perplexity happen in the last layers}, and this trend is the same for both BERT and RoBERTa. Interestingly, we observe no difference between CoLA and SST-2 fine-tuning in this experiment.

\subsection{Discussion}
\label{sec:discussion}

In the following, we discuss the main implications of our experiments and analysis. 

\begin{enumerate}
    \item We conclude that fine-tuning indeed does affect the representations of a pre-trained model and in particular those of the last hidden layers, which is supported by our perplexity analysis. However, our perplexity analysis does not reveal whether these changes have a positive or negative effect on the encoding of linguistic knowledge.
    \item Some fine-tuning/probing task combinations result in substantial improvements in probing accuracy when using CLS-pooling. 
    Our attention analysis supports our interpretation that the improvement in probing accuracy can not simply be attributed to the encoding of linguistic knowledge, but can at least partially be explained by changes in the attention distribution for the CLS token.
    We note that this is also consistent with our findings that the improvement in probing accuracy vanishes when comparing to the mean-pooling baseline.
    \item Some other task combinations have a negative effect on the probing task performance, suggesting that the linguistic knowledge our probing classifiers are testing for is indeed no longer (linearly) accessible. However, it remains unclear whether fine-tuning indeed removes the linguistic knowledge our probing classifiers are testing for from the representations or whether it is simply no longer linearly separable. 
    We are planning to further investigate this in future work.
\end{enumerate}

\section{Conclusion}

We investigated the interplay between fine-tuning and layer-wise sentence-level probing accuracy and found that fine-tuning can lead to substantial changes in probing accuracy. However, these changes vary greatly depending on the encoder model and fine-tuning and probing task combination. Our analysis of attention distributions after fine-tuning showed, that changes in probing accuracy can not be attributed to the encoding of linguistic knowledge alone but might as well be caused by changes in the attention distribution. At the same time, our perplexity analysis showed that fine-tuning has profound effects on the representations of a pre-trained model but our probing analysis can not sufficiently detail whether it leads to forgetting of the probed linguistic information. Hence we argue that the effects of fine-tuning on pre-trained representations should be carefully interpreted.

\section*{Acknowledgments}

We thank Badr Abdullah for his comments and suggestions. We would also like to thank the reviewers for their useful comments and feedback, in particular R1. This work was funded by the Deutsche Forschungsgemeinschaft (DFG, German Research Foundation) – project-id 232722074 – SFB 1102.

\bibliography{anthology,emnlp2020}
\bibliographystyle{acl_natbib}

\appendix
\section{Appendices}
\label{sec:appendix}

\section{Hyperparameters and Task Statistics}

Table \ref{tab:hyperparams} shows hyperparamters used when fine-tuning BERT, RoBERTa, and ALBERT on CoLA, SST-2, RTE, and SQuAD. On SST-2 training for a single epoch was sufficient and we didn't observe a significant improvement when training for more epochs.

\begin{table}[h]
    \centering
    \begin{tabular}{ll} 
        \toprule
        \textbf{Hyperparameter} & \textbf{Value} \\
        \midrule 
        Learning rate  & $\num{2e-05}$   \\
        Warmup steps & $10\%$ \\ 
        Learning rate schedule  & warmup-constant  \\
        Batch size & 32 \\
        Epochs & 3 (1 for SST-2)  \\
        Weight decay & $0.01$ \\
        Dropout & $0.1$ \\
        Attention dropout & $0.1$ \\
        Classifier dropout & $0.1$ \\
        Adam $\epsilon$ & $\num{1e-08}$ \\
        Adam $\beta_1$ & $0.9$ \\
        Adam $\beta_2$ & $0.99$ \\
        Max. gradient norm & $1.0$ \\
        \bottomrule
    \end{tabular}
    \caption{Hyperparamters used when fine-tuning.}
    \label{tab:hyperparams}
\end{table}

\noindent Table \ref{tab:task-stats} shows number of training and development samples for each of the fine-tuning datasets considered in our experiments. Additionally, we report the metric used to evaluate performance for each of the tasks.

\begin{table}[h]
    \centering
    \begin{tabular}{lcccccc} 
        \toprule
        \multirow{2}{*}{\textbf{Statistics}} & \multicolumn{4}{c}{\textbf{Task}}\\ \cmidrule{2-5}
         & CoLA & SST-2 & RTE & SQuAD  \\
        \midrule 
        training &  8.6k & 67k & 2.5 & 87k \\
        validation & 1,043 & 874  & 278 & 10k \\ 
        metric & MCC & Acc. & Acc. & EM/$F_1$ \\
        \bottomrule
    \end{tabular}
    \caption{Fine-tuning task statistics.}
    \label{tab:task-stats}
\end{table}

\section{Additional Results}

Table \ref{tab:aggregated-acc-deltas-cls-mean-rte-squad11} shows the effect of fine-tuning on RTE and SQuAD on the layer-wise accuracy for all three encoder models across the three probing tasks. 

\begin{table*}[ht!]
    \centering
        \begin{subfigure}[t]{\textwidth}
            \centering
                \begin{tabular}{lrrrrrrrr} 
                    \toprule
                    \multirow{4}{*}{\textbf{Probing Task}} &  \multicolumn{8}{c}{\textbf{\textcolor{bert-orange}{BERT-base-cased}}}\\ \cmidrule{2-9}
                    & \multicolumn{4}{c}{CLS-pooling} & \multicolumn{4}{c}{mean-pooling}  \\ \cmidrule{2-9}
                    & \multicolumn{2}{c}{RTE} & \multicolumn{2}{c}{SQuAD} & \multicolumn{2}{c}{RTE} & \multicolumn{2}{c}{SQuAD}  \\ 
                     &  0 -- 6 & 7 -- 12 &  0 -- 6 & 7 -- 12 & 0 -- 6 & 7 -- 12 & 0 -- 6 & 7 -- 12 \\
                    \midrule
                    bigram-shift &  $-0.21$ & $-0.39$ & $-0.05$ & $-1.50$ & $-0.07$ & $-0.31$ & $-0.54$ & $-1.66$  \\
                    coordinate-inversion &  $-0.43$ & $-0.36$ & $0.04$ & $0.56$ & $0.05$ & $0.13$ & $-0.03$ & $0.10$  \\
                    odd-man-out &  $0.09$ & $0.38$ & $-0.21$ & $-1.89$ & $0.09$ & $0.01$ & $-0.28$ & $-1.73$ \\
                \end{tabular}
        \end{subfigure}\vspace{.5em}%
        \\
        \begin{subfigure}[t]{\textwidth}
            \centering
                \begin{tabular}{lrrrrrrrr} 
                        \toprule
                        \multirow{4}{*}{\textbf{Probing Task}} &  \multicolumn{8}{c}{\textbf{\textcolor{roberta-blue}{RoBERTa-base}}}\\ \cmidrule{2-9}
                        & \multicolumn{4}{c}{CLS-pooling} & \multicolumn{4}{c}{mean-pooling}  \\ \cmidrule{2-9}
                    & \multicolumn{2}{c}{RTE} & \multicolumn{2}{c}{SQuAD} & \multicolumn{2}{c}{RTE} & \multicolumn{2}{c}{SQuAD}  \\ 
                         &  0 -- 6 & 7 -- 12 &  0 -- 6 & 7 -- 12 & 0 -- 6 & 7 -- 12 & 0 -- 6 & 7 -- 12  \\
                        \midrule
                        bigram-shift &  $-0.51$ & $0.44$ & $-1.17$ & $-4.33$ & $-0.09$ & $-1.32$ & $-0.28$ & $-3.09$ \\
                    coordinate-inversion &  $-0.35$ & $3.27$ & $0.29$ & $0.50$ & $0.30$ & $-0.48$ & $0.20$ & $0.05$  \\
                    odd-man-out &  $-0.11$ & $1.22$ & $-0.76$ & $-3.01$ & $-0.04$ & $-1.96$ & $-0.21$ & $-3.58$ \\
                    \end{tabular}
        \end{subfigure}\vspace{.5em}%
        \\
        \begin{subfigure}[t]{\textwidth}
            \centering
                \begin{tabular}{lrrrrrrrr} 
                    \toprule
                    \multirow{4}{*}{\textbf{Probing Task}} &  \multicolumn{8}{c}{\textbf{\textcolor{albert-green}{ALBERT-base-v1}}}\\ \cmidrule{2-9}
                    & \multicolumn{4}{c}{CLS-pooling} & \multicolumn{4}{c}{mean-pooling}  \\ \cmidrule{2-9}
                    & \multicolumn{2}{c}{RTE} & \multicolumn{2}{c}{SQuAD} & \multicolumn{2}{c}{RTE} & \multicolumn{2}{c}{SQuAD}  \\ 
                     &  0 -- 6 & 7 -- 12 &  0 -- 6 & 7 -- 12 & 0 -- 6 & 7 -- 12 & 0 -- 6 & 7 -- 12  \\
                    \midrule
                    bigram-shift &  $0.29$ & $-0.43$ & $-0.38$ & $-3.46$ & $-0.13$ & $-0.82$ & $-0.60$ & $-3.11$ \\
                    coordinate-inversion &  $0.46$ & $-0.44$ & $0.32$ & $0.92$ & $0.13$ & $-0.38$ & $0.04$ & $-0.27$  \\
                    odd-man-out &  $-0.03$ & $0.17$ & $-0.65$ & $-2.91$ & $-0.17$ & $-0.85$ & $-0.55$ & $-3.18$ \\
                    \bottomrule
                \end{tabular}
        \end{subfigure}%
    \caption{Change in probing accuracy $\Delta$ (in $\%$) of \textbf{RTE} and \textbf{SQuAD} fine-tuned models compared to the pre-trained models when using CLS and mean-pooling. We average the difference in probing accuracy over two different layers groups: layers 0 to 6 and layers 7 to 12.}
    \label{tab:aggregated-acc-deltas-cls-mean-rte-squad11}
\end{table*}

\begin{figure*}[h!]
    \centering
    
    \begin{subfigure}[t]{.33\textwidth}
        \centering
        \includegraphics[width=1.0\textwidth]{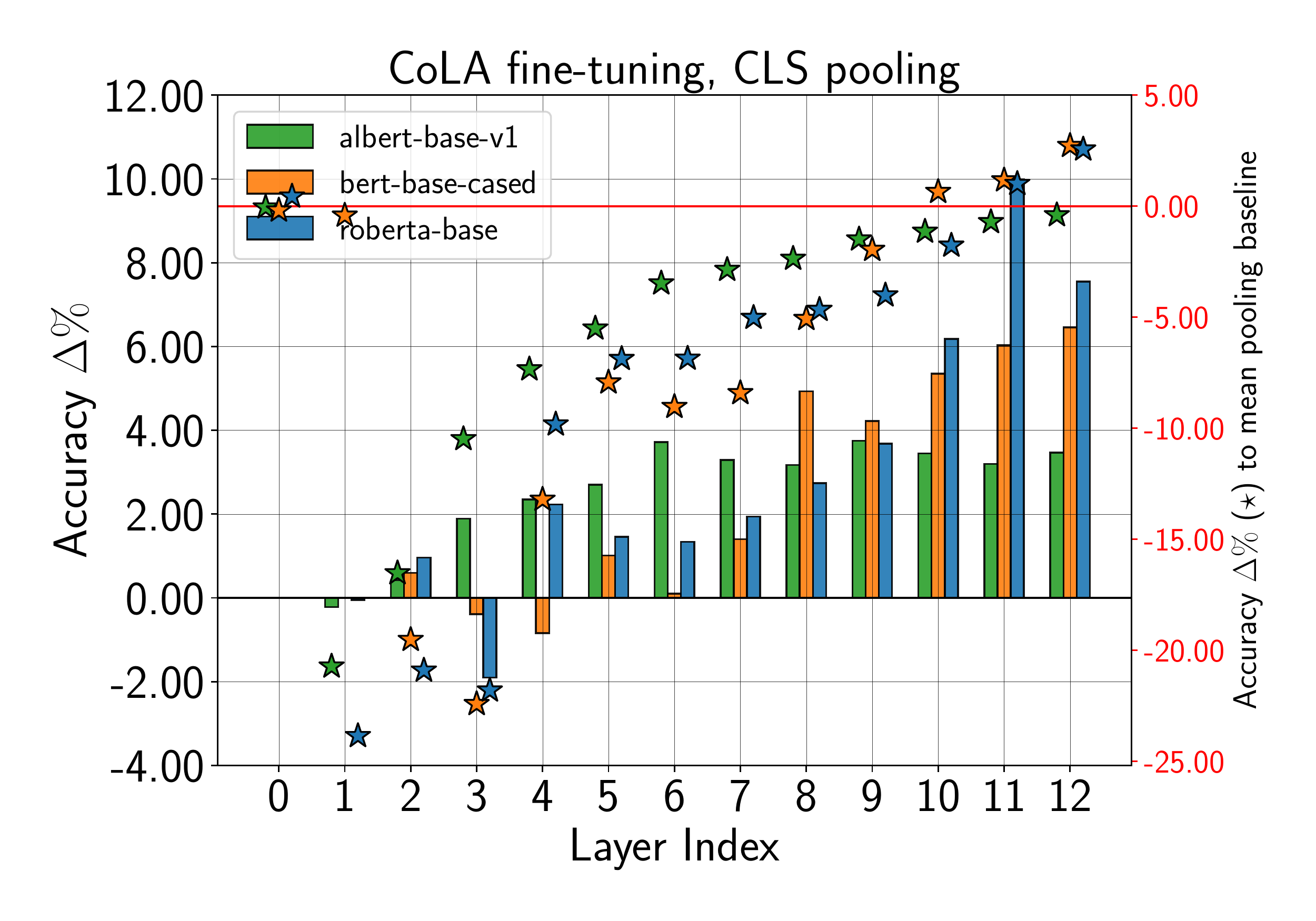}
        \caption{bigram-shift}
    \end{subfigure}%
    ~
    \begin{subfigure}[t]{.33\textwidth}
        \centering
        \includegraphics[width=1.0\textwidth]{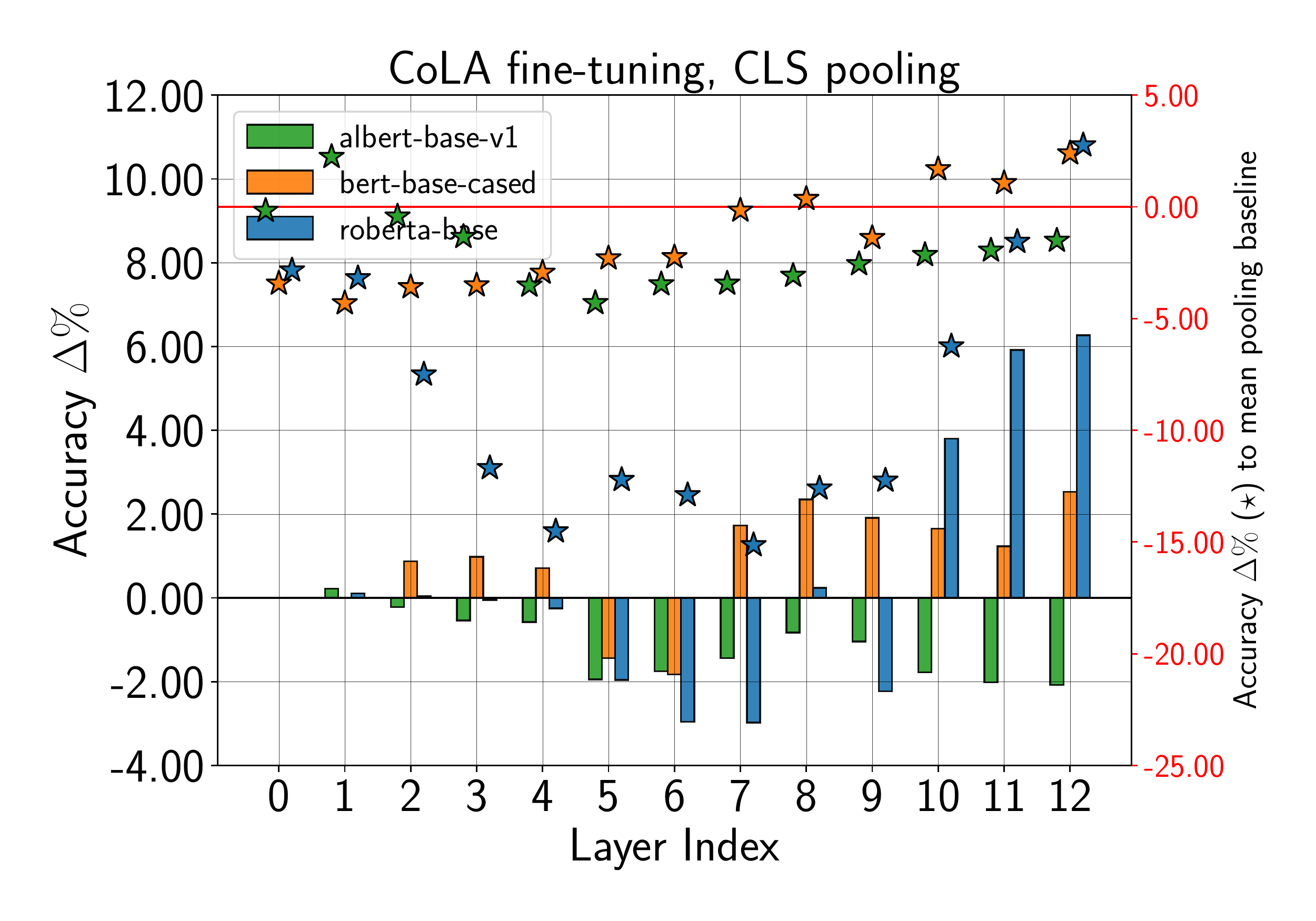}
        \caption{coordination-inversion}
    \end{subfigure}%
    ~ 
    \begin{subfigure}[t]{.33\textwidth}
        \centering
        \includegraphics[width=1.0\textwidth]{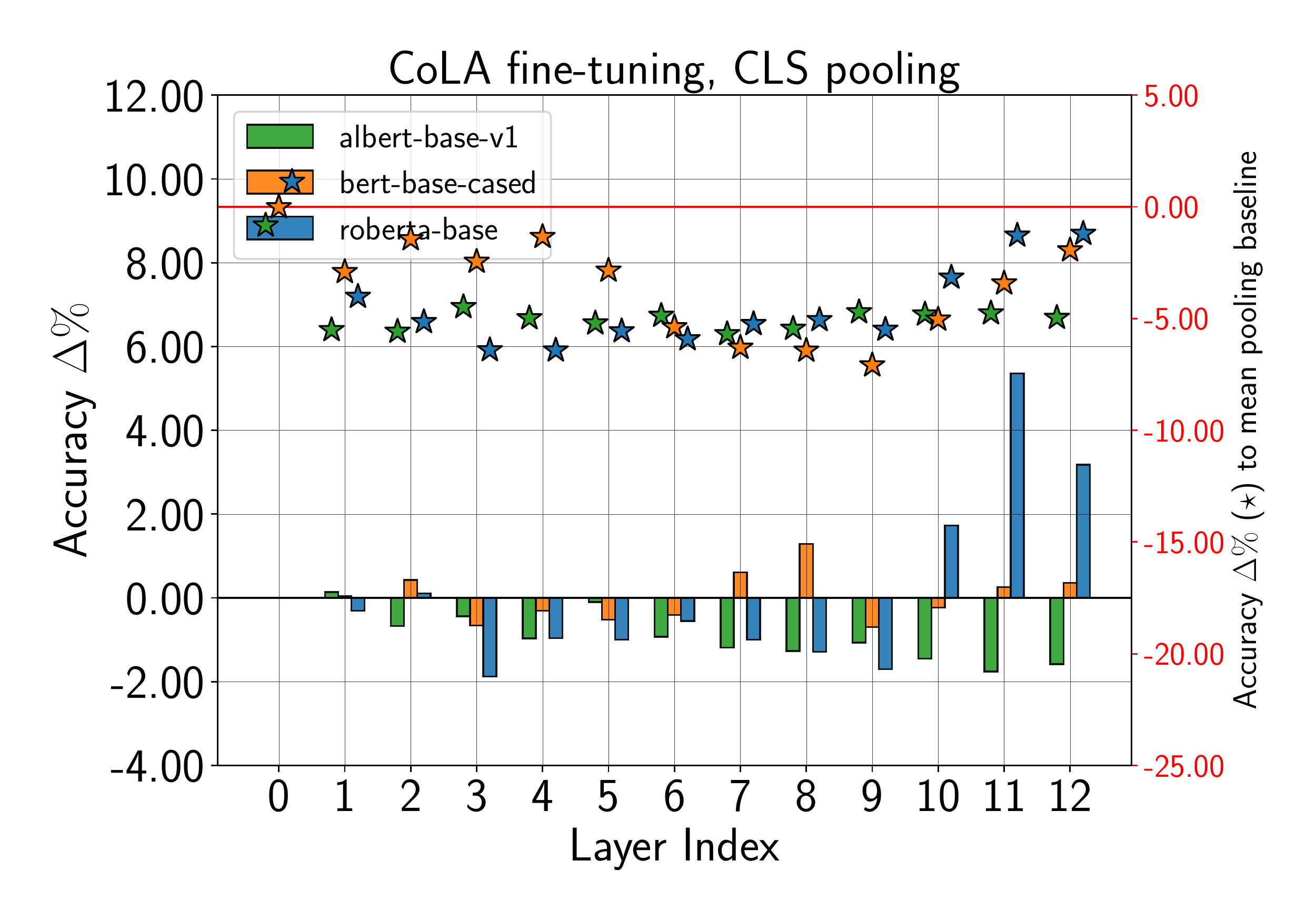}
        \caption{odd-man-out}
    \end{subfigure}%
    \ 
    \begin{subfigure}[t]{.33\textwidth}
        \centering
        \includegraphics[width=1.0\textwidth]{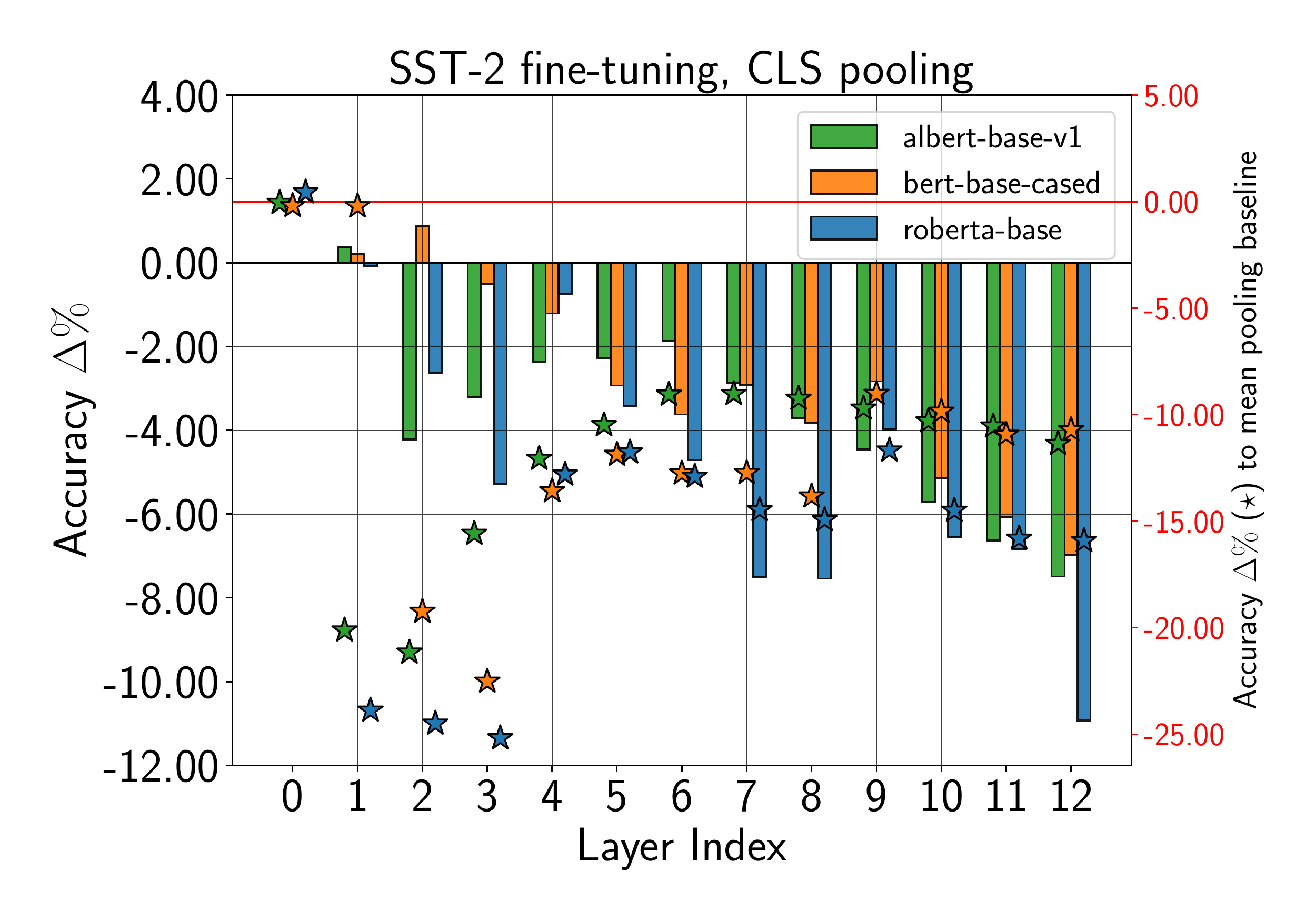}
        \caption{bigram-shift}
    \end{subfigure}%
    ~
    \begin{subfigure}[t]{.33\textwidth}
        \centering
        \includegraphics[width=1.0\textwidth]{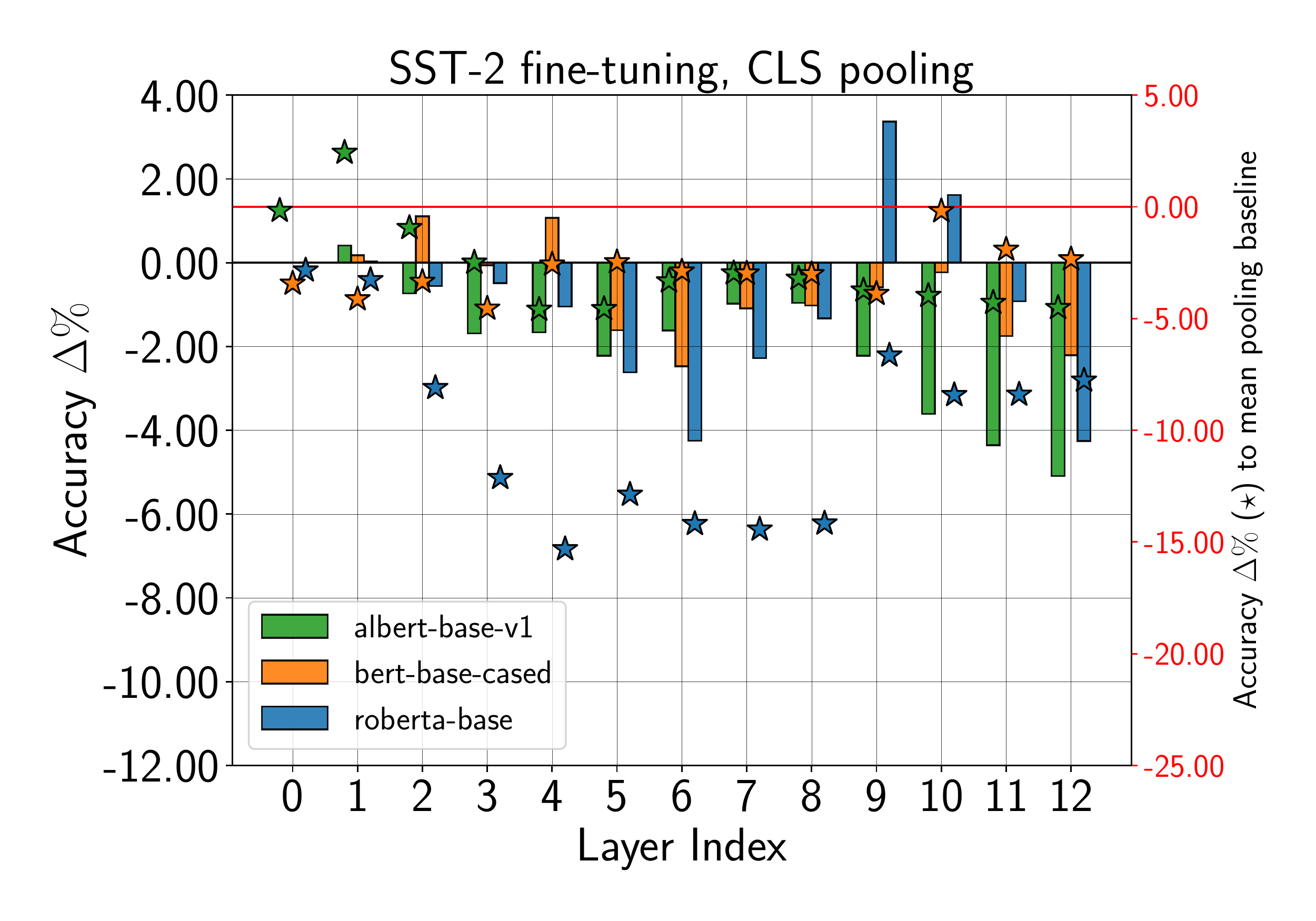}
        \caption{coordination-inversion}
    \end{subfigure}%
    ~ 
    \begin{subfigure}[t]{.33\textwidth}
        \centering
        \includegraphics[width=1.0\textwidth]{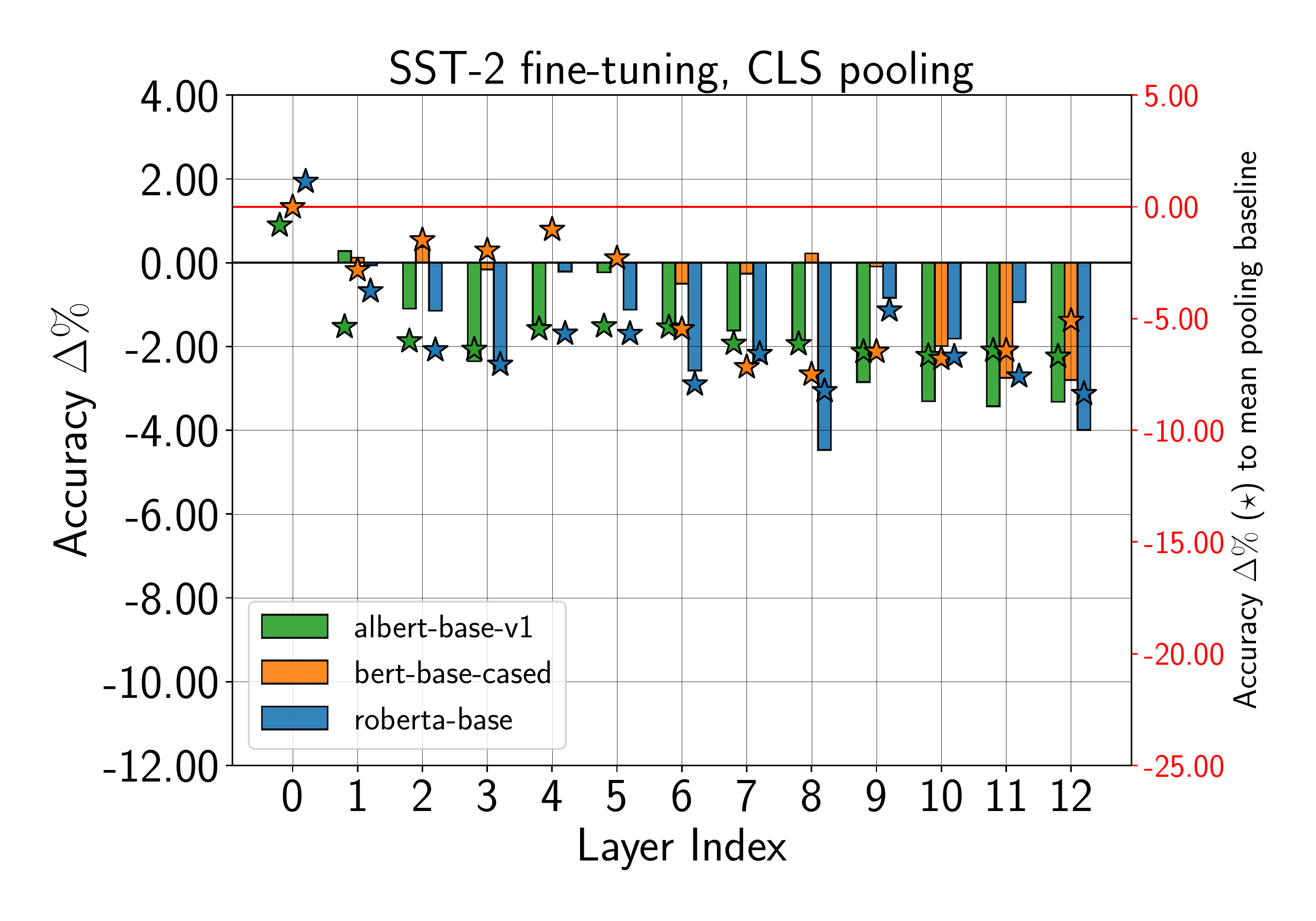}
        \caption{odd-man-out}
    \end{subfigure}%
    \ 
    \begin{subfigure}[t]{.33\textwidth}
        \centering
        \includegraphics[width=1.0\textwidth]{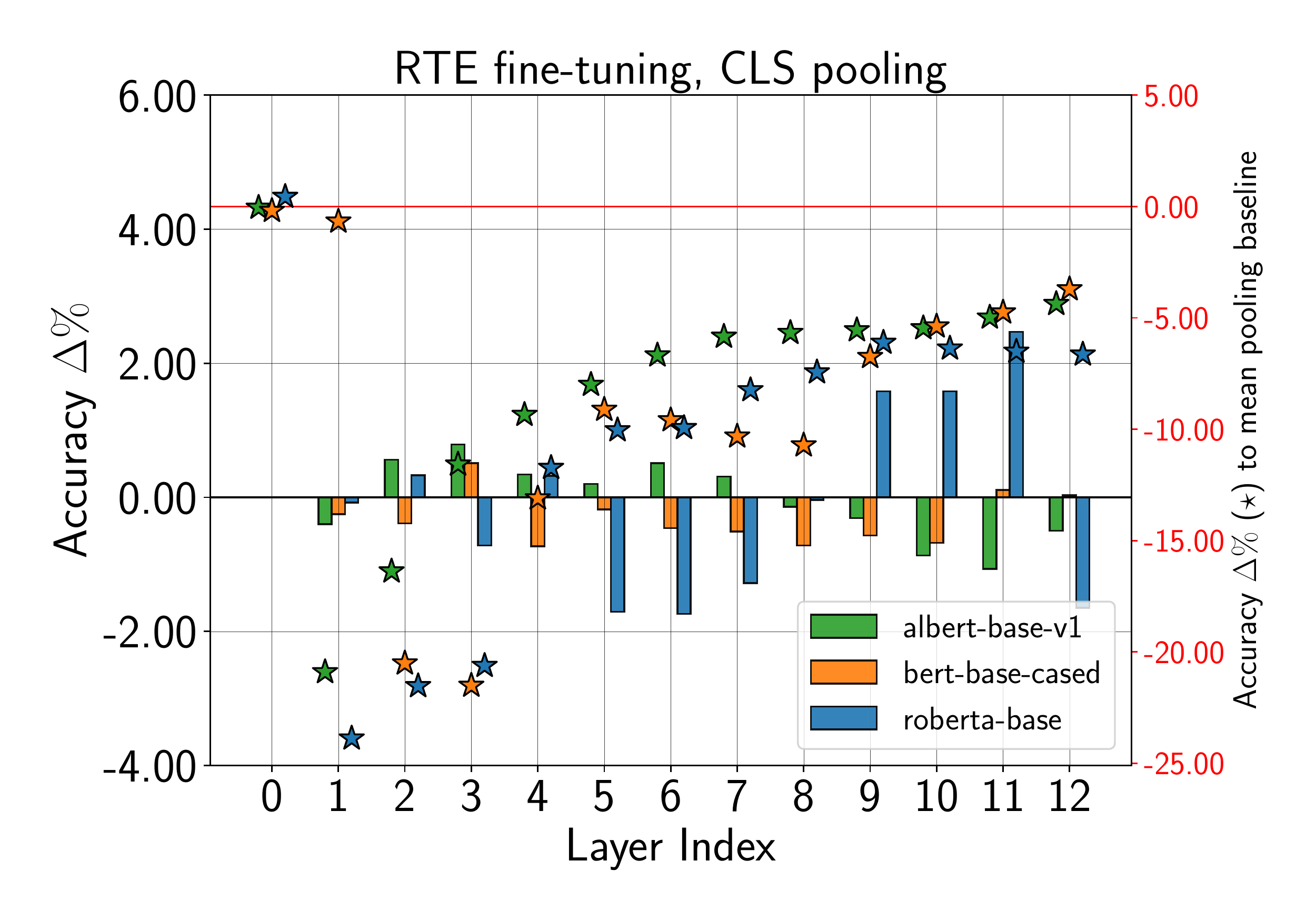}
        \caption{bigram-shift}
    \end{subfigure}%
    ~
    \begin{subfigure}[t]{.33\textwidth}
        \centering
        \includegraphics[width=1.0\textwidth]{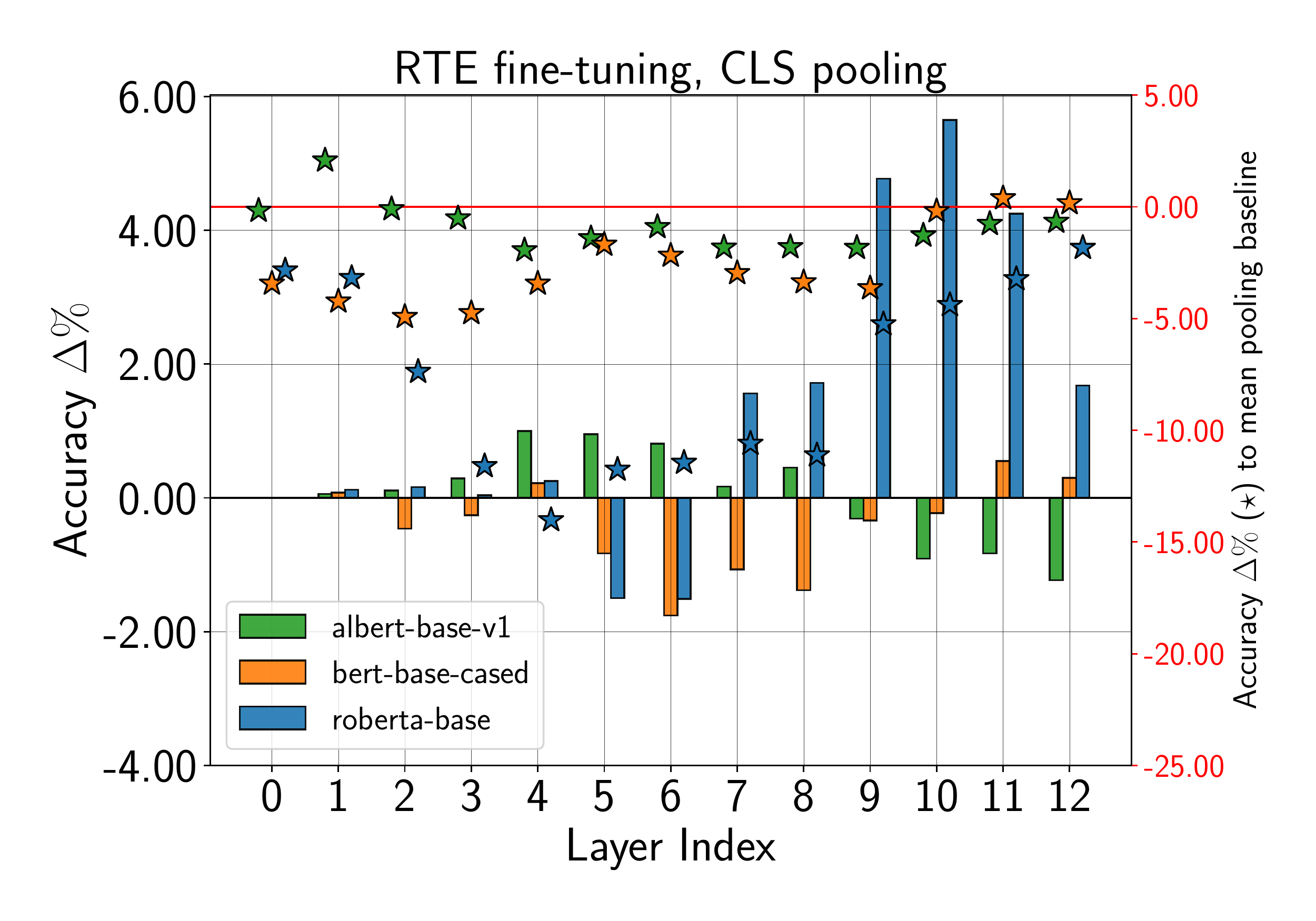}
        \caption{coordination-inversion}
    \end{subfigure}%
    ~ 
    \begin{subfigure}[t]{.33\textwidth}
        \centering
        \includegraphics[width=1.0\textwidth]{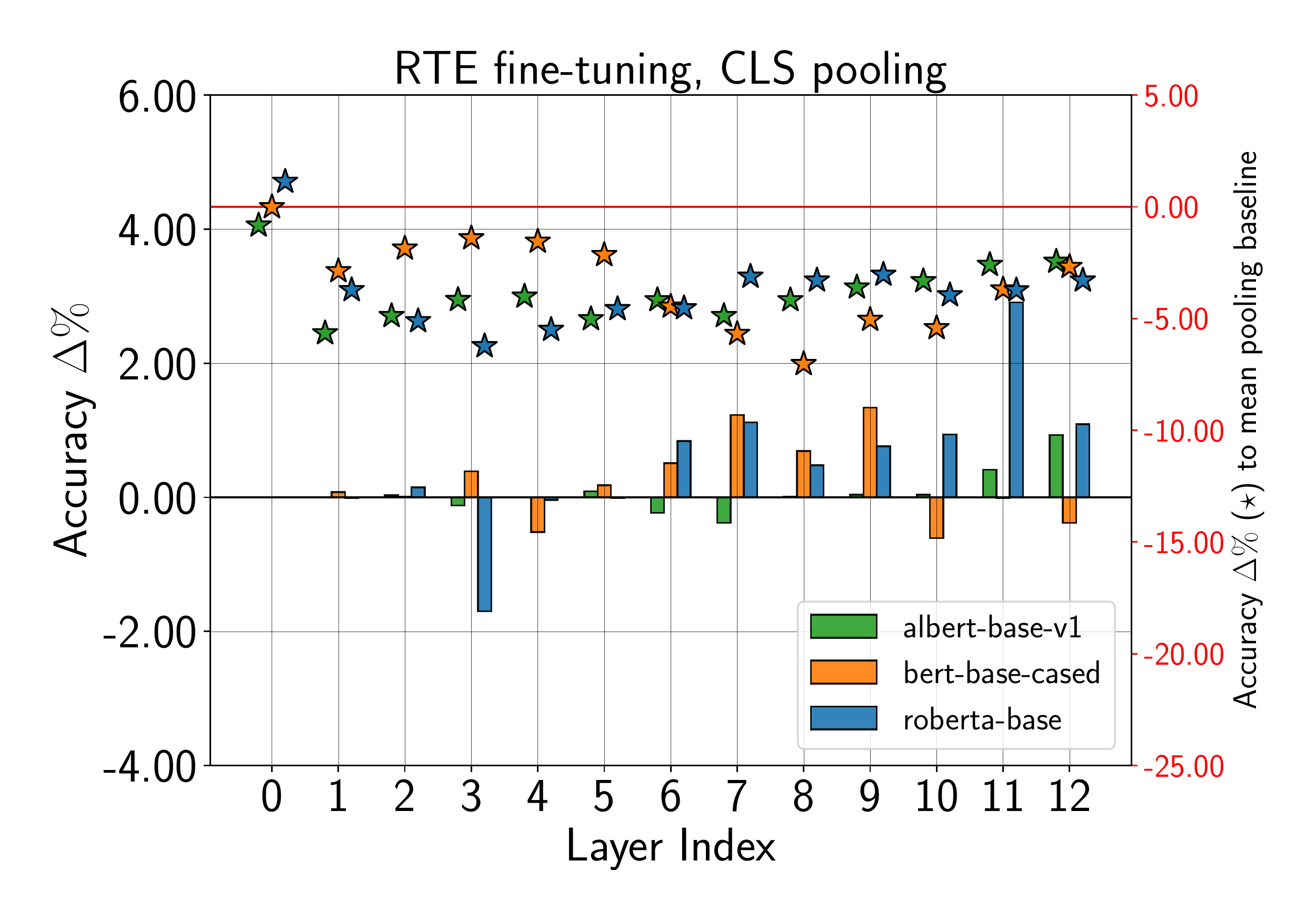}
        \caption{odd-man-out}
    \end{subfigure}%
    \ 
    \begin{subfigure}[t]{.33\textwidth}
        \centering
        \includegraphics[width=1.0\textwidth]{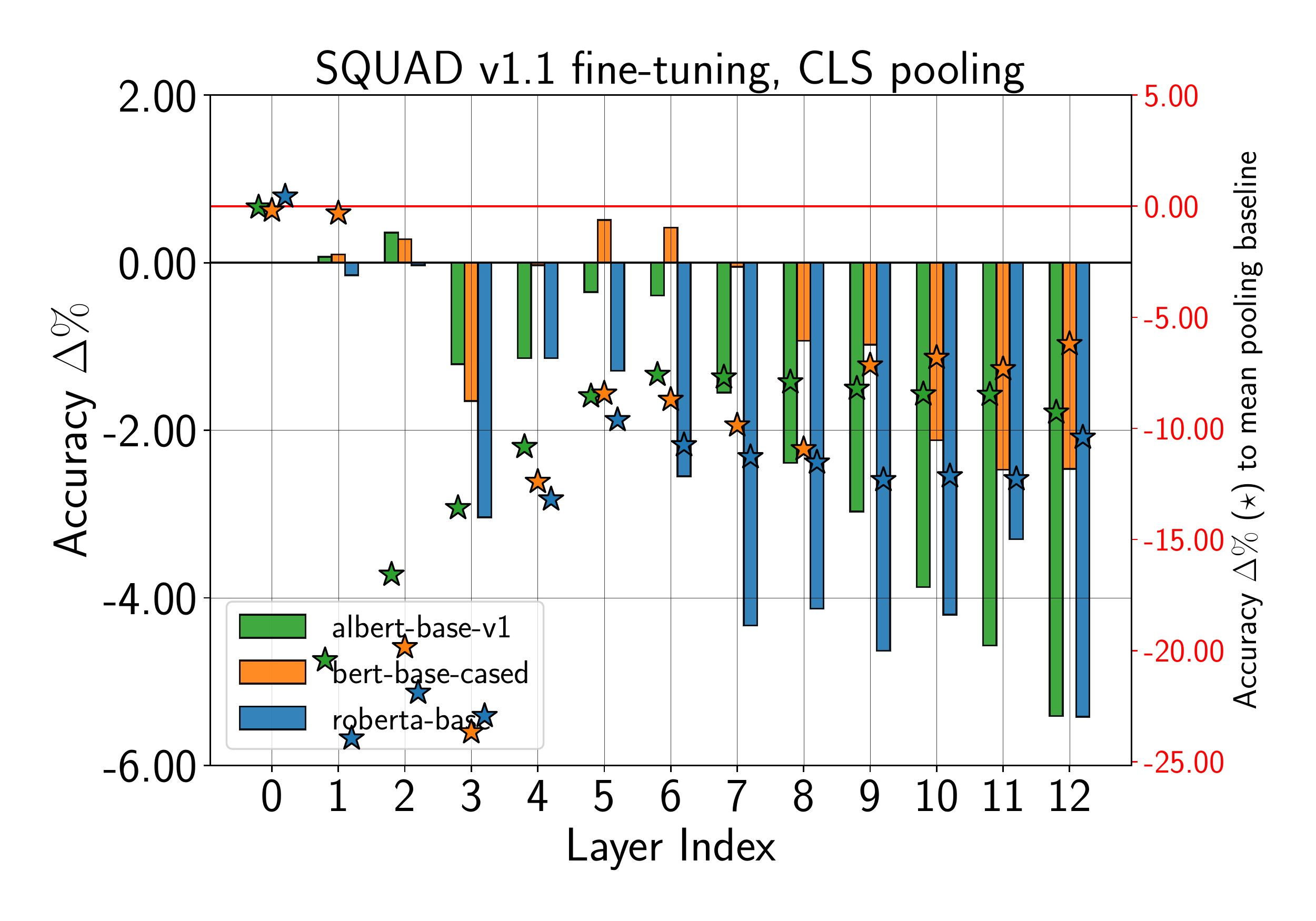}
        \caption{bigram-shift}
    \end{subfigure}%
    ~
    \begin{subfigure}[t]{.33\textwidth}
        \centering
        \includegraphics[width=1.0\textwidth]{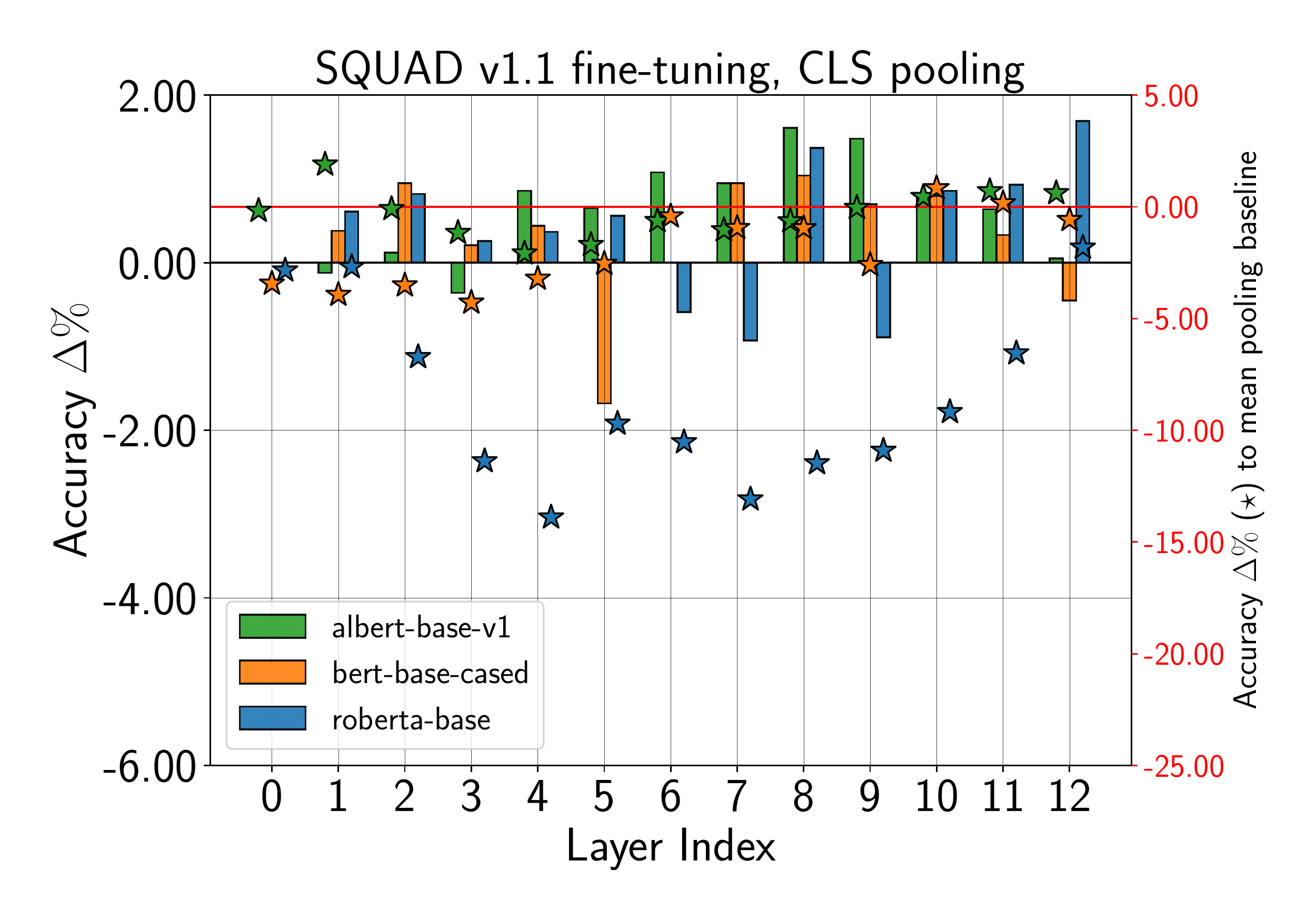}
        \caption{coordination-inversion}
    \end{subfigure}%
    ~ 
    \begin{subfigure}[t]{.33\textwidth}
        \centering
        \includegraphics[width=1.0\textwidth]{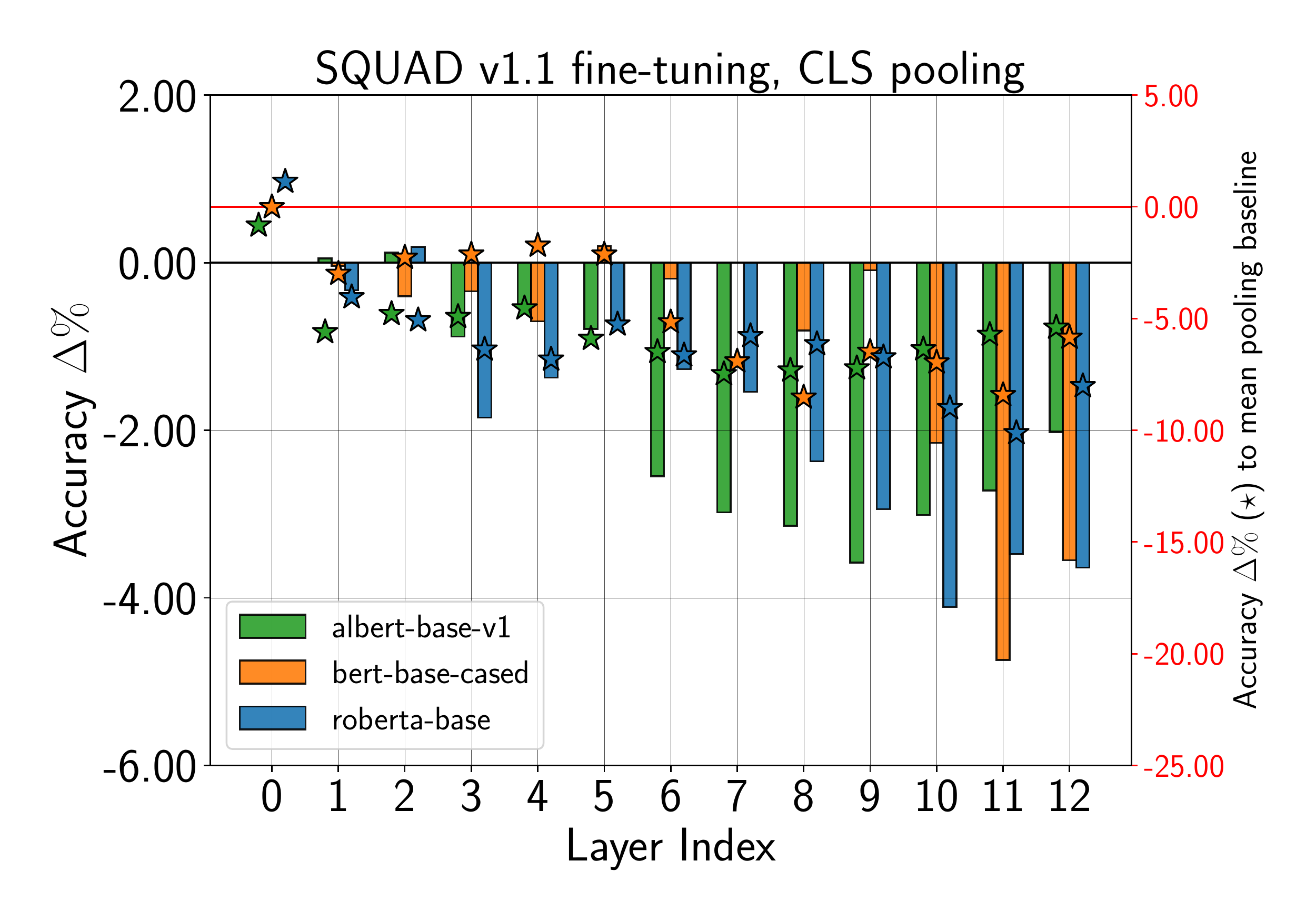}
        \caption{odd-man-out}
    \end{subfigure}%
    
    \caption{Difference in probing accuracy $\Delta$ (in $\%$) when using CLS-pooling after fine-tuning on \textbf{CoLA}, \textbf{SST-2}, \textbf{RTE}, and \textbf{SQuAD} for all three encoder models BERT, RoBERTa, and ALBERT across all probing taks considered in this work. The second y-axis shows layer-wise improvement over the mean-pooling baselines (stars) on the respective task.}
    \label{fig:aggregated-acc-deltas-cola-and-stars}
\end{figure*}

\begin{figure*}[h!]
    \centering
    \begin{subfigure}[t]{.33\textwidth}
        \centering
        \includegraphics[width=1.0\textwidth]{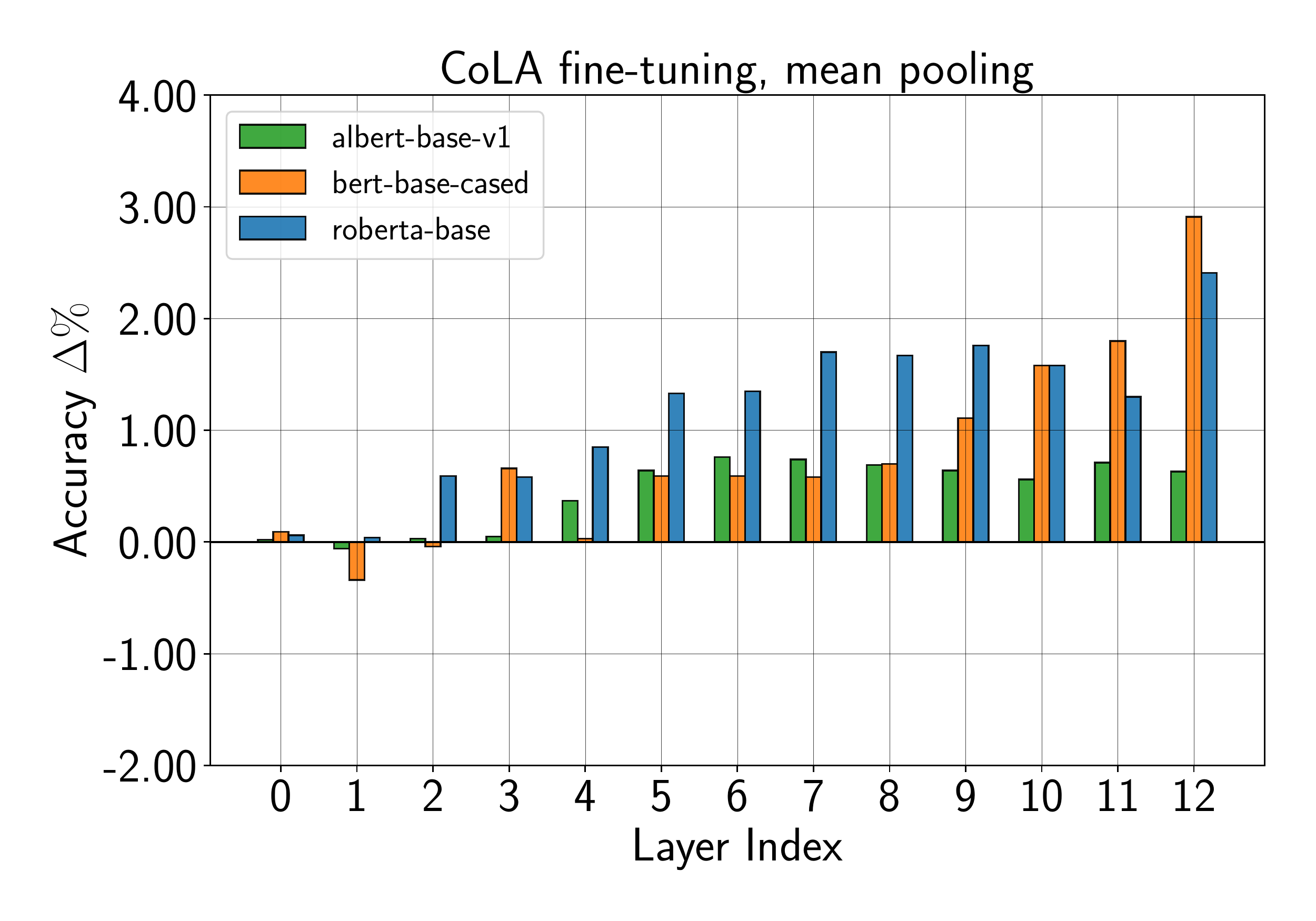}
        \caption{bigram-shift}
    \end{subfigure}%
    ~
    \begin{subfigure}[t]{.33\textwidth}
        \centering
        \includegraphics[width=1.0\textwidth]{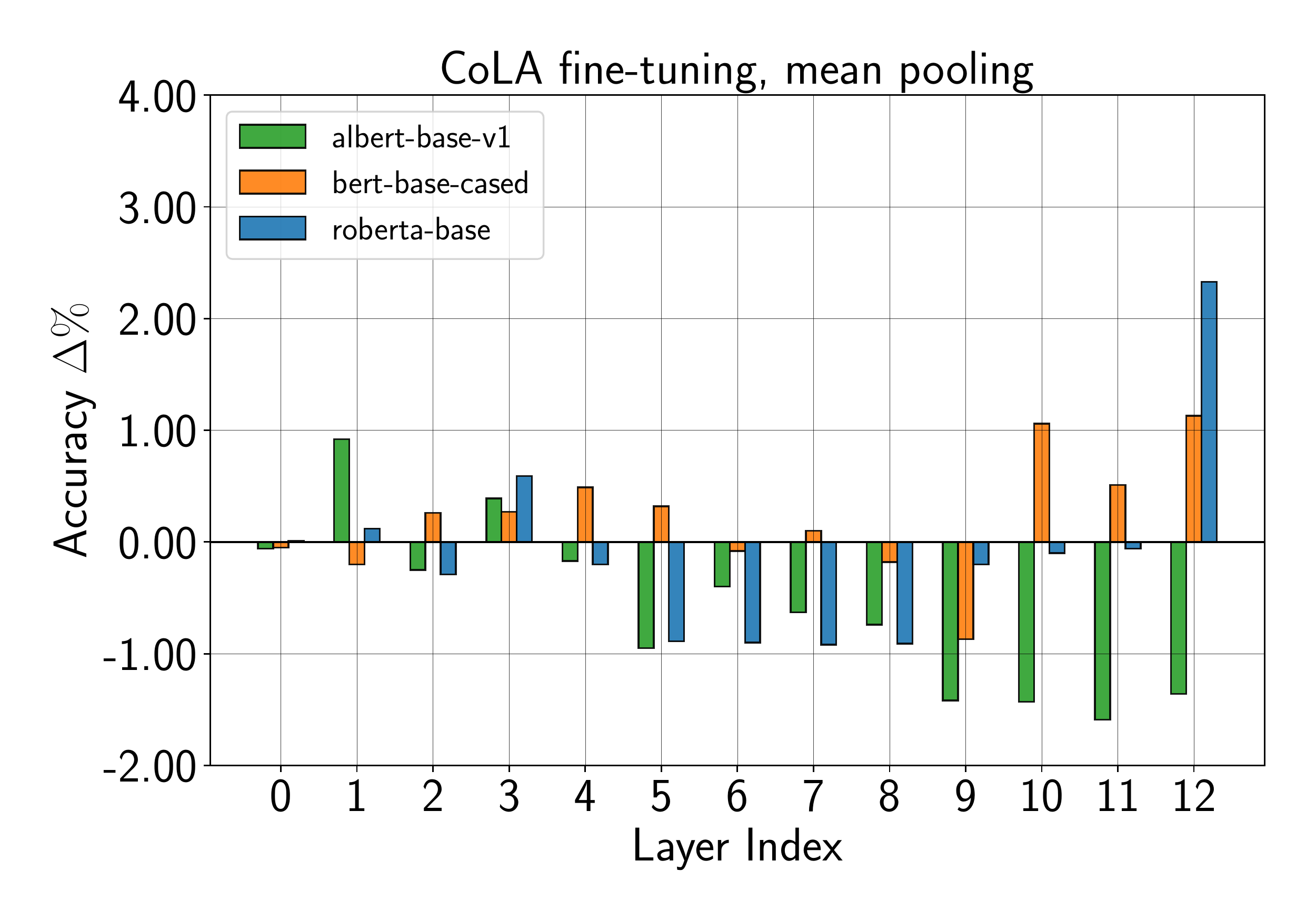}
        \caption{coordination-inversion}
    \end{subfigure}%
    ~ 
    \begin{subfigure}[t]{.33\textwidth}
        \centering
        \includegraphics[width=1.0\textwidth]{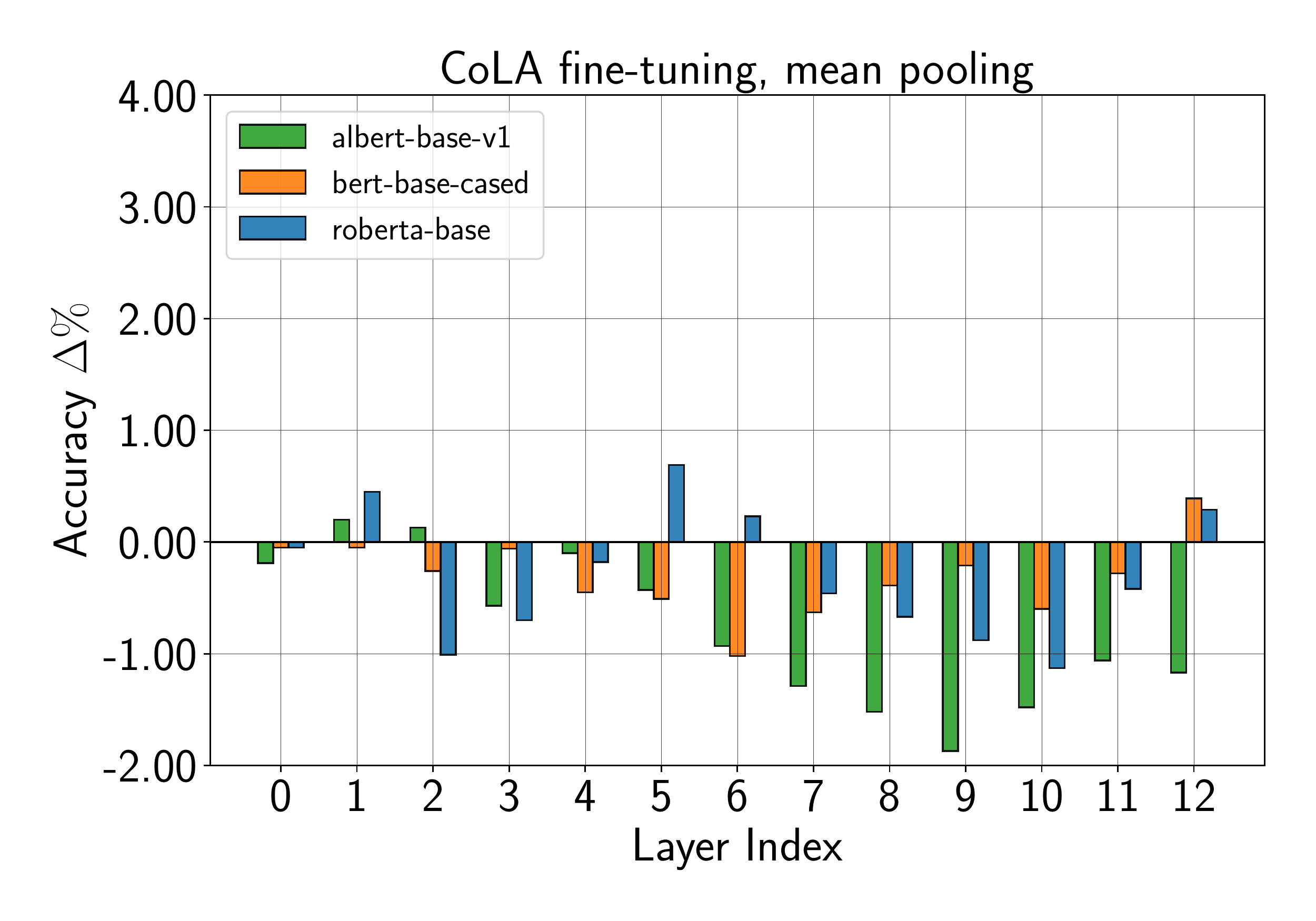}
        \caption{odd-man-out}
    \end{subfigure}%
        \ 
    \begin{subfigure}[t]{.33\textwidth}
        \centering
        \includegraphics[width=1.0\textwidth]{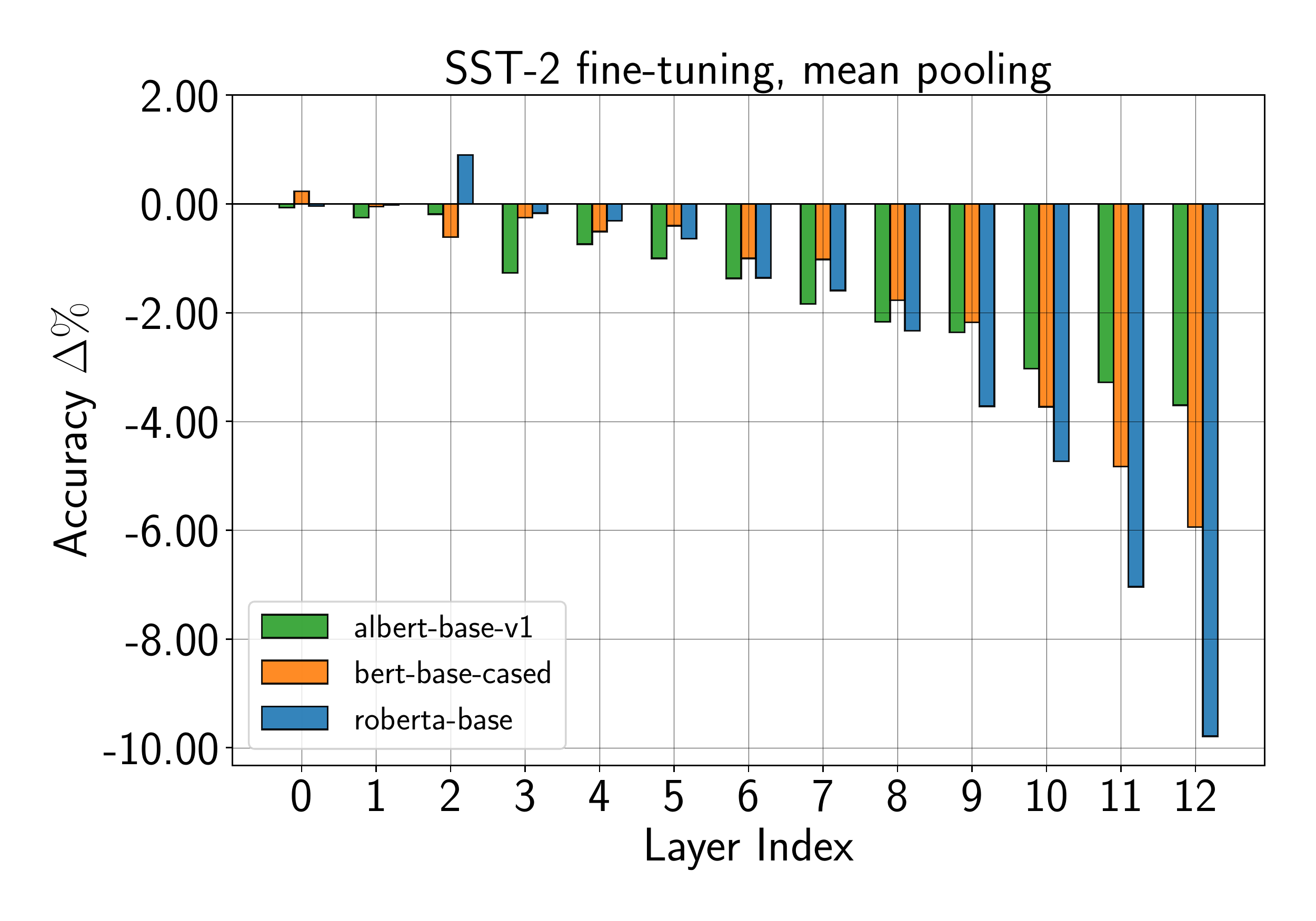}
        \caption{bigram-shift}
    \end{subfigure}%
    ~
    \begin{subfigure}[t]{.33\textwidth}
        \centering
        \includegraphics[width=1.0\textwidth]{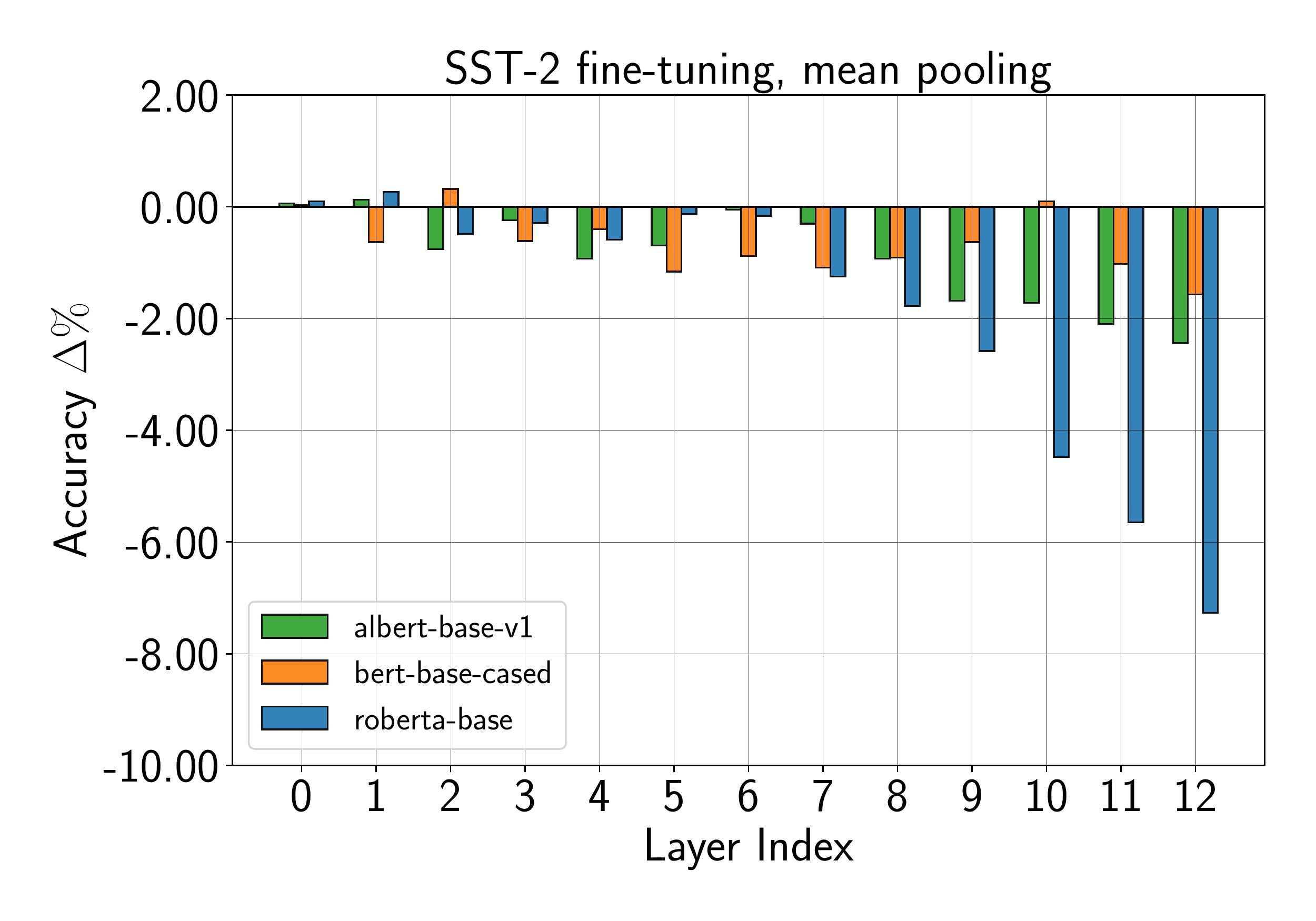}
        \caption{coordination-inversion}
    \end{subfigure}%
    ~ 
    \begin{subfigure}[t]{.33\textwidth}
        \centering
        \includegraphics[width=1.0\textwidth]{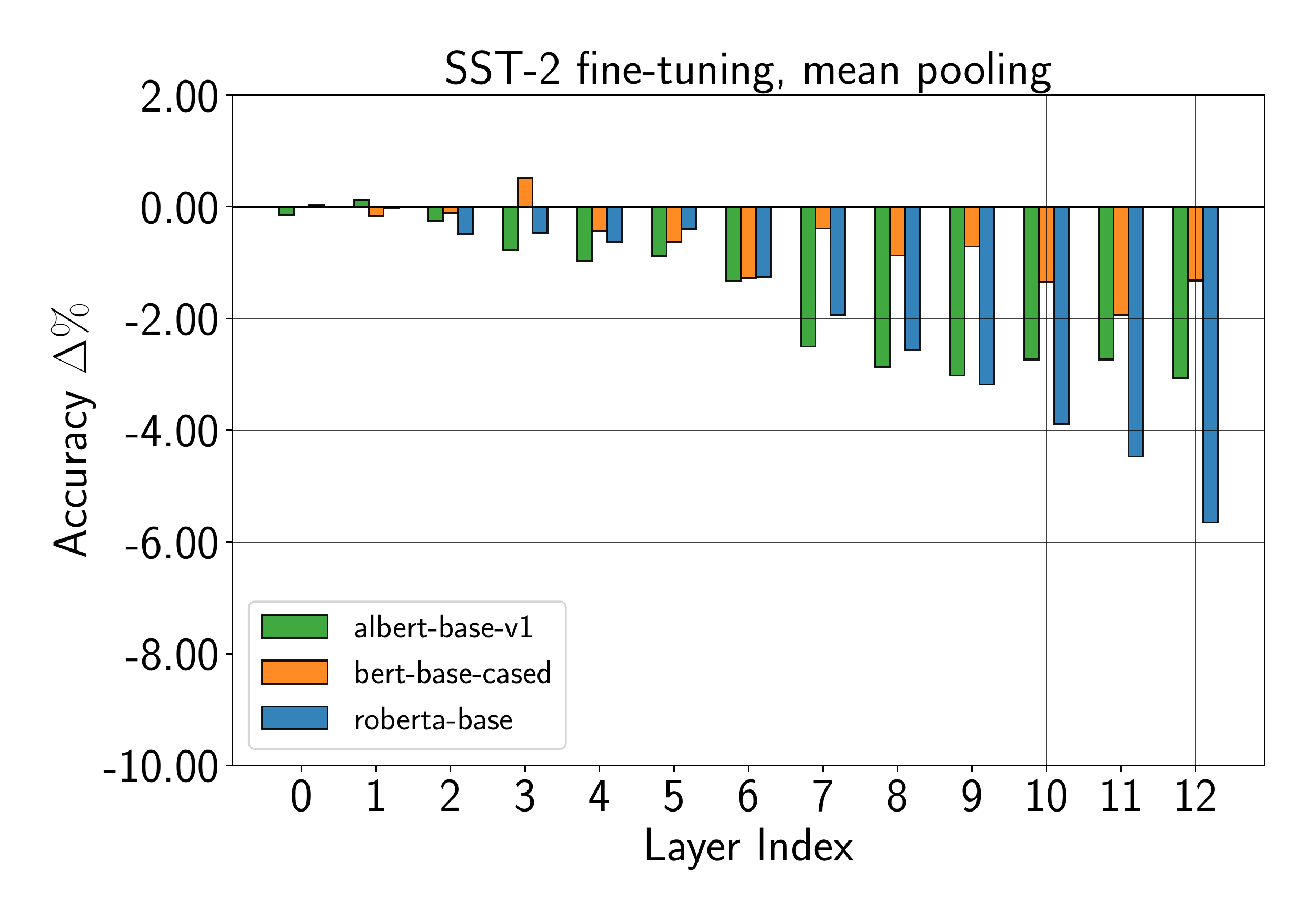}
        \caption{odd-man-out}
    \end{subfigure}%
    \
    \begin{subfigure}[t]{.33\textwidth}
        \centering
        \includegraphics[width=1.0\textwidth]{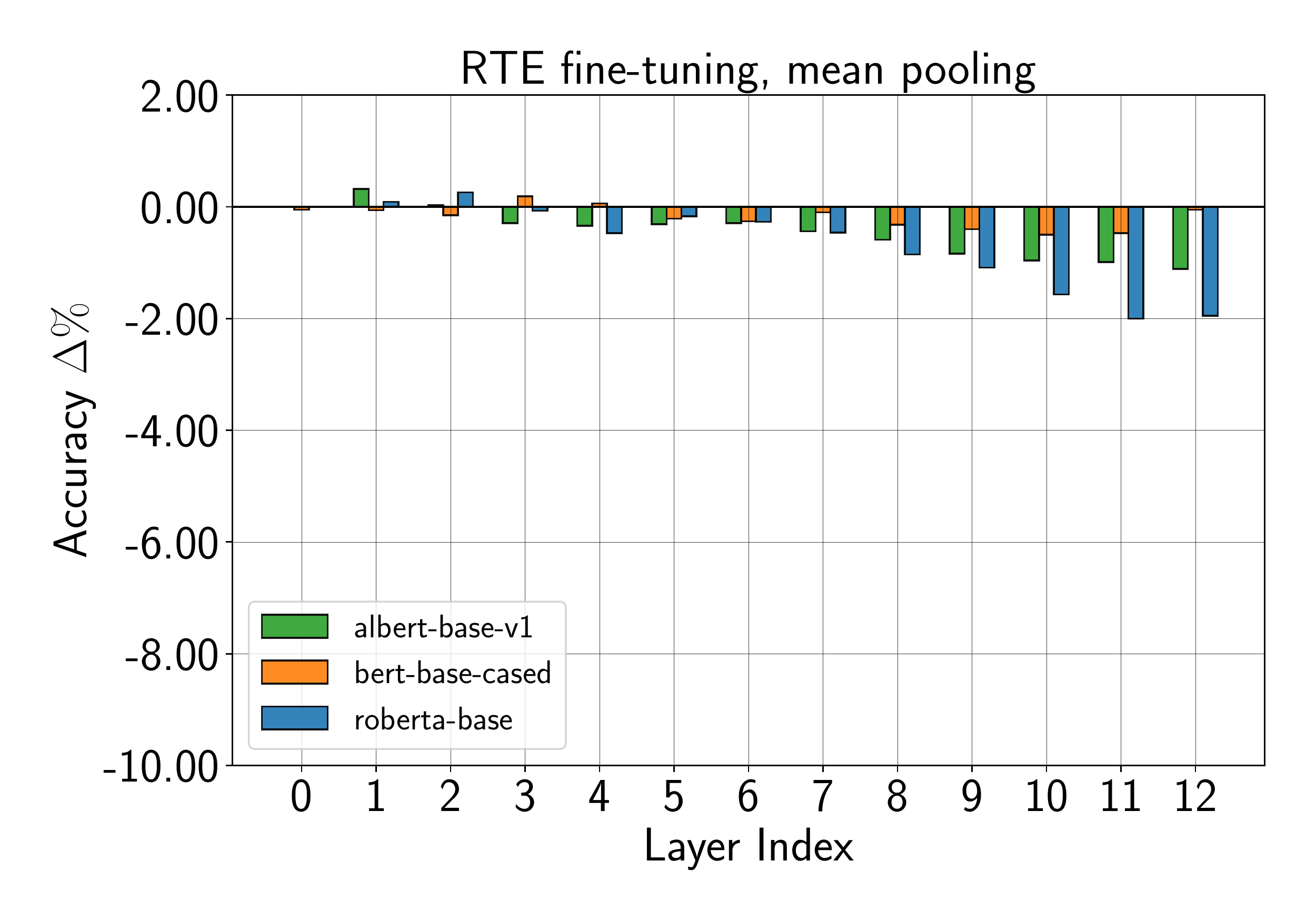}
        \caption{bigram-shift}
    \end{subfigure}%
    ~
    \begin{subfigure}[t]{.33\textwidth}
        \centering
        \includegraphics[width=1.0\textwidth]{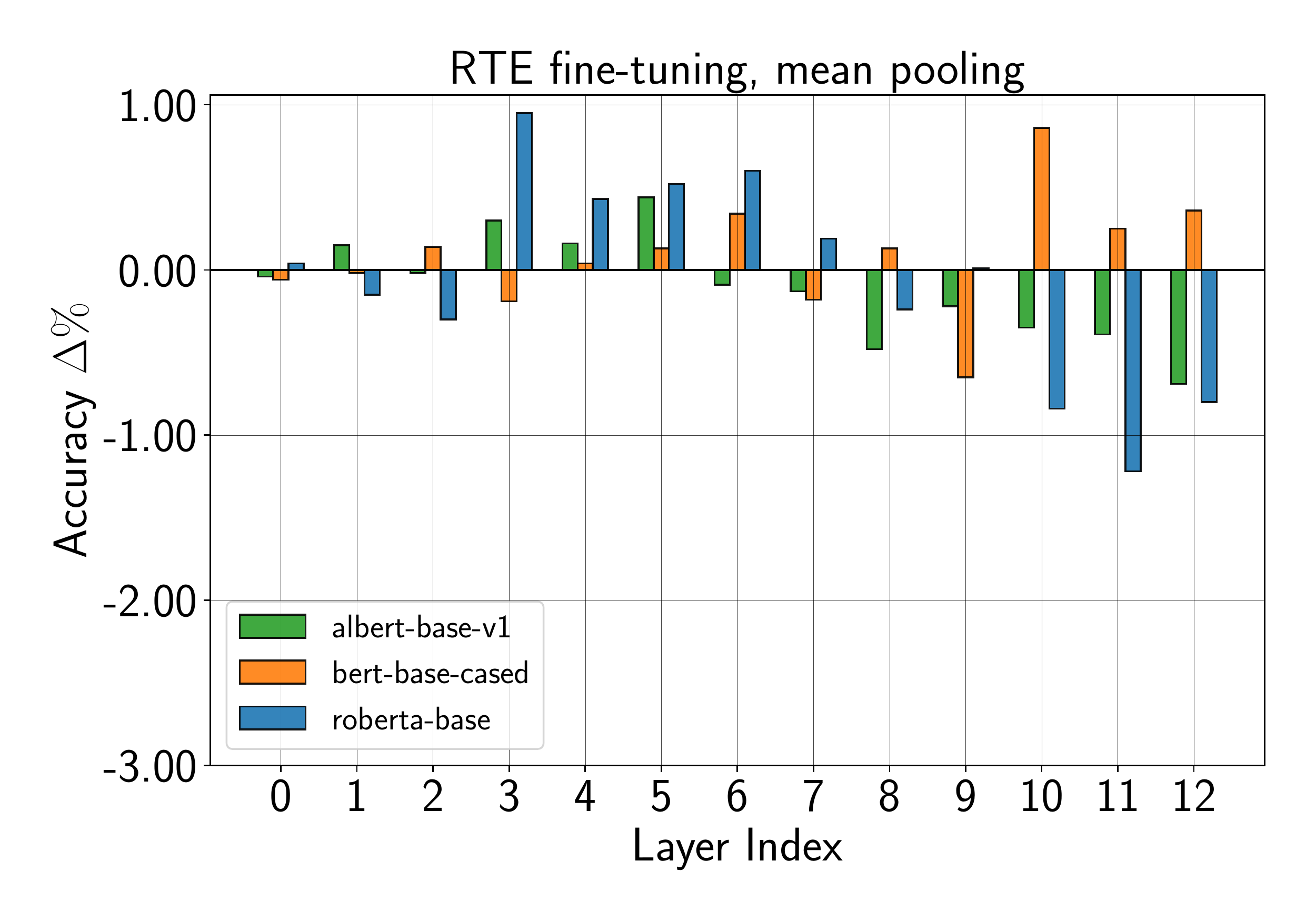}
        \caption{coordination-inversion}
    \end{subfigure}%
    ~ 
    \begin{subfigure}[t]{.33\textwidth}
        \centering
        \includegraphics[width=1.0\textwidth]{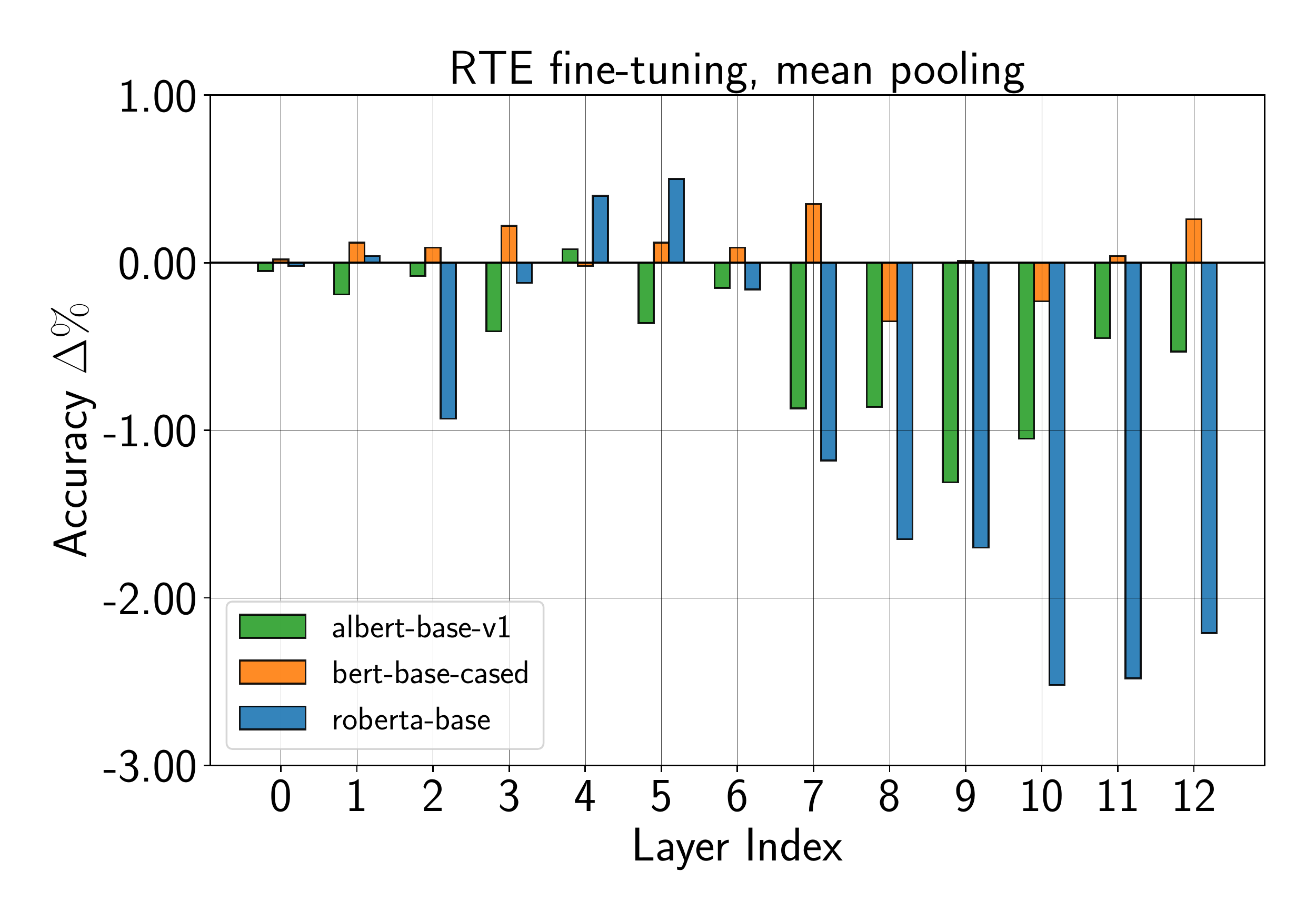}
        \caption{odd-man-out}
    \end{subfigure}%
    \ 
    \begin{subfigure}[t]{.33\textwidth}
        \centering
        \includegraphics[width=1.0\textwidth]{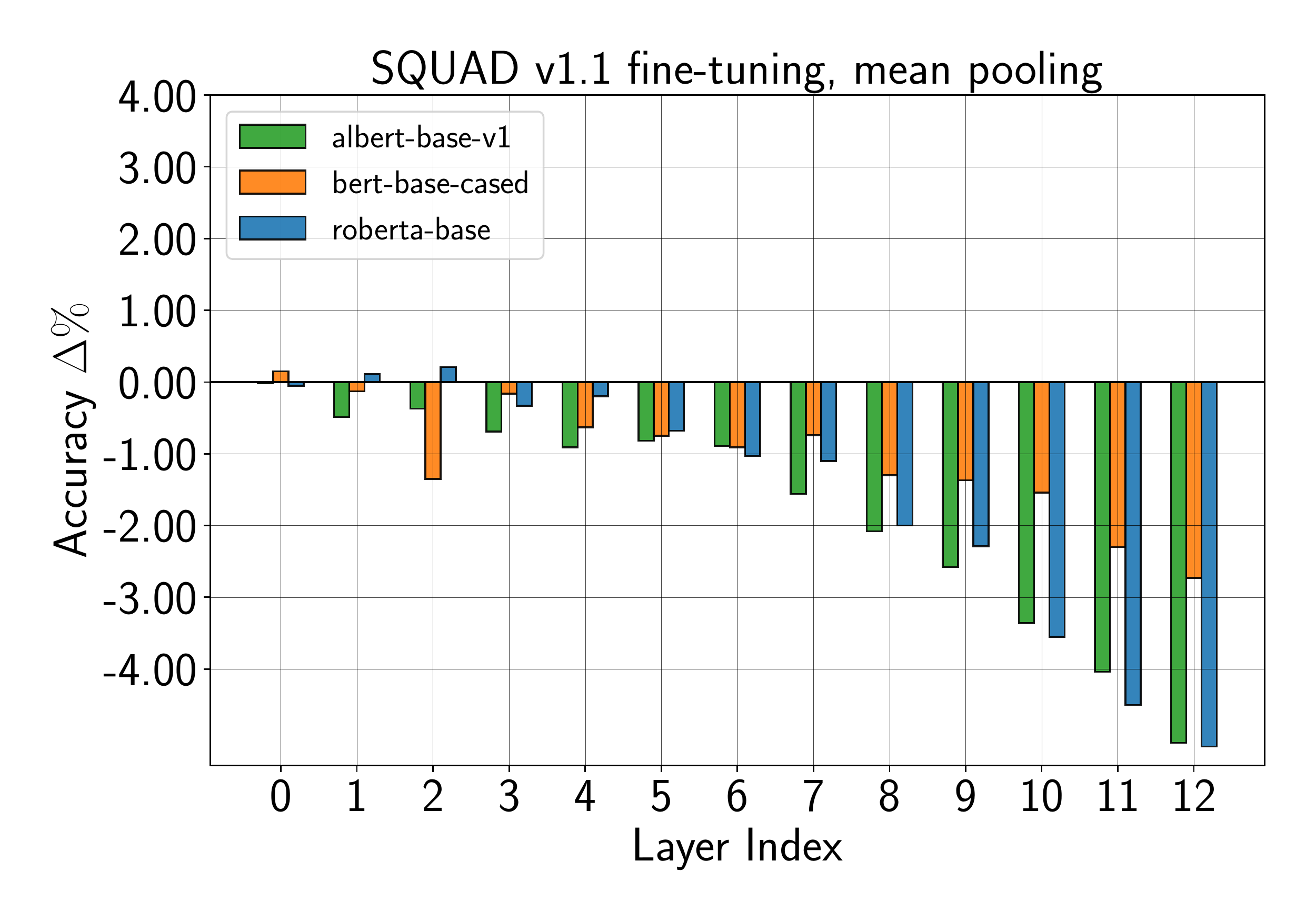}
        \caption{bigram-shift}
    \end{subfigure}%
    ~
    \begin{subfigure}[t]{.33\textwidth}
        \centering
        \includegraphics[width=1.0\textwidth]{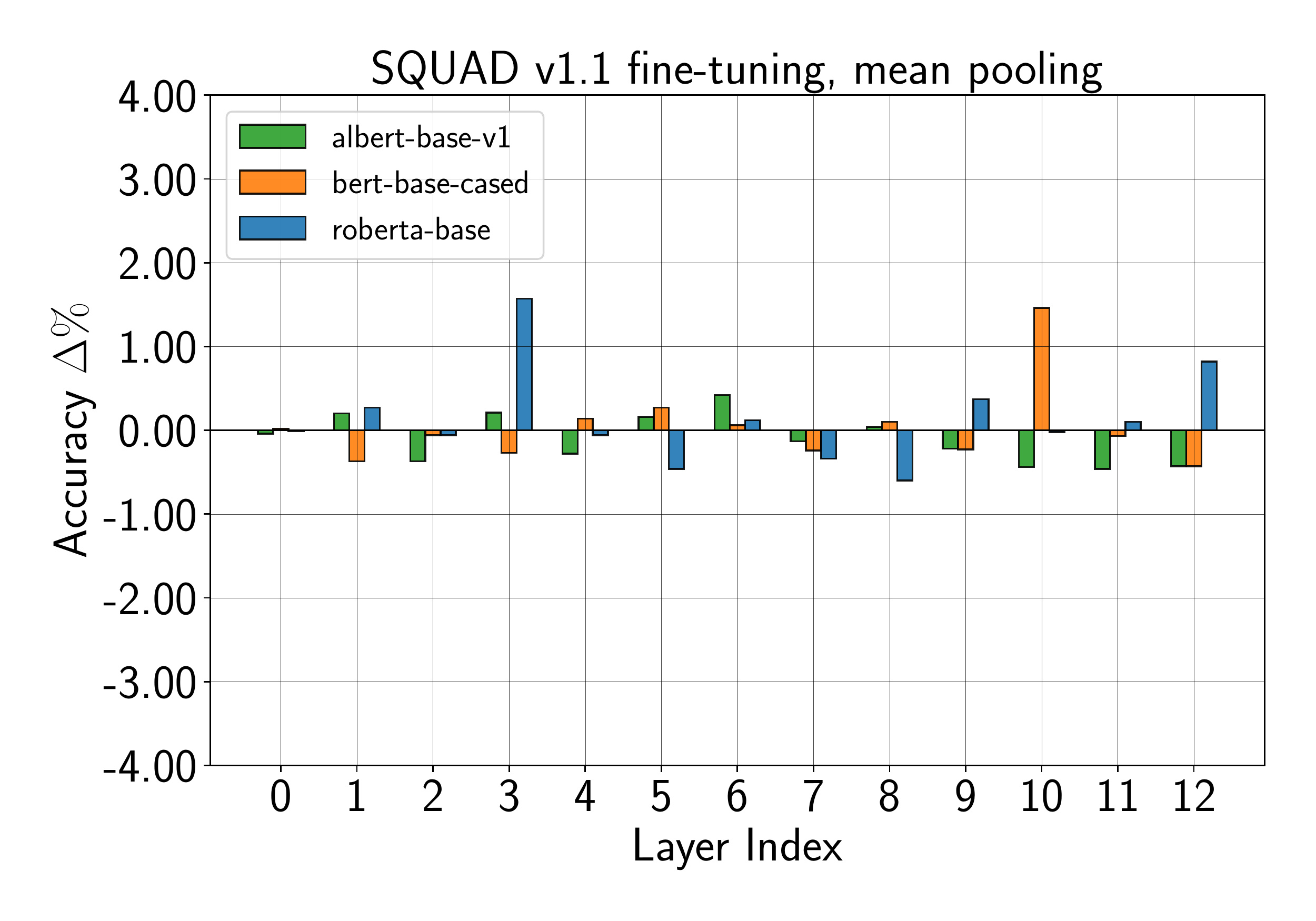}
        \caption{coordination-inversion}
    \end{subfigure}%
    ~ 
    \begin{subfigure}[t]{.33\textwidth}
        \centering
        \includegraphics[width=1.0\textwidth]{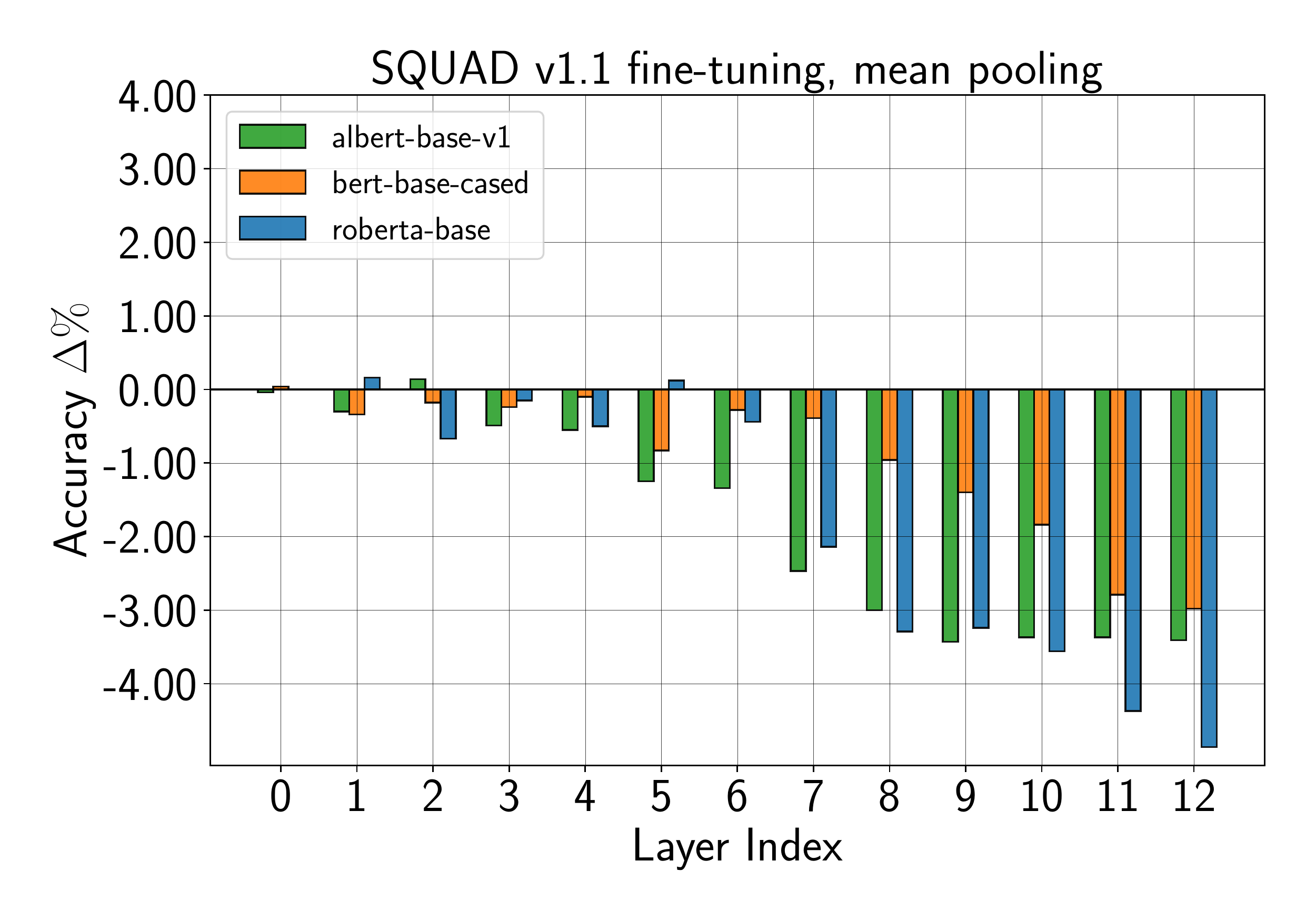}
        \caption{odd-man-out}
    \end{subfigure}%

    \caption{Difference in probing accuracy $\Delta$ (in $\%$) when using mean-pooling after fine-tuning on \textbf{CoLA}, \textbf{SST-2}, \textbf{RTE}, and \textbf{SQuAD} for all three encoder models BERT, RoBERTa, and ALBERT across all probing tasks considered in this work.}
    
    \label{fig:full-mean-cola}

\end{figure*}

Figure \ref{fig:aggregated-acc-deltas-cola-and-stars} and Figure \ref{fig:full-mean-cola} show the change in probing accuracy $\Delta$ (in $\%$) across all probing tasks when fine-tuning on CoLA, SST-2, RTE, and SQuAD using CLS-pooling and mean-pooling, respectively. The second y-axis in Figure \ref{fig:aggregated-acc-deltas-cola-and-stars} shows the layer-wise difference after fine-tuning compared to the mean-pooling baseline. Note that only in very few cases this differences is larger than zero.x

\end{document}